\documentclass[11pt]{article}

\usepackage{acl}

\usepackage{times}
\usepackage{latexsym}
\usepackage{amsmath}
\usepackage{amssymb}
\usepackage{booktabs}
\usepackage{multirow}
\usepackage{xcolor}
\usepackage[T1]{fontenc}
\usepackage[utf8]{inputenc}
\usepackage{microtype}
\usepackage{inconsolata}
\usepackage{graphicx}

\graphicspath{{figures/}{figures_link/}}

\title{CircuitLens: Reasoning Circuits as Data Selection Signals for Reinforcement Learning with Verifiable Rewards}

\author{Zhuofan Chen\textsuperscript{1}\quad
        Ziqian Jiao\textsuperscript{1}\quad
        Yikai Cui\textsuperscript{1}\quad
        Zhixin Cai\textsuperscript{1}\quad \\
        {\bf Jun Bai\textsuperscript{2}}\quad
        {\bf Wenge Rong\textsuperscript{1}}\\
  \textsuperscript{1}School of Computer Science and Engineering, Beihang University, Beijing, China\\
  \textsuperscript{2}Beijing Institute for General Artificial Intelligence, China\\
  \texttt{\{zhuofanchen, jiaoziqian, cuiyikai, caizhixin, w.rong\}@buaa.edu.cn},\\
  \texttt{baijun@bigai.ai}
}

\begin{document}
\maketitle

% ============================================================
% ABSTRACT
% ============================================================

\begin{abstract}
Reinforcement learning with verifiable rewards (RLVR) is sensitive to which problems a model trains on, yet existing selection criteria---difficulty filtering, hand-curation, reward-trajectory scoring---assess data value as an intrinsic property of problems, independent of the model that will learn from them.
We introduce Circuit Reasoning Score (CRS), a selection signal derived from 46 reasoning-sensitive attention heads identified via contrastive ablation, computed in a single forward pass on the frozen base model without reward labels or rollouts.
CRS runs against the intuitive hypothesis that stronger reasoning-circuit engagement produces better training data: on Qwen2.5-Math-7B, the lowest-engagement decile improves over random selection on three medium-difficulty benchmarks (GSM8K $+2.0$\,pp, OlympiadBench $+1.6$\,pp, Minerva $+2.9$\,pp), while the highest-engagement decile gains less and is indistinguishable from the middle decile.
The advantage has boundary conditions:
on a domain-curated pool no selection method separates from the others;
at 1.5B scale the useful direction differs;
and the lowest-reward training condition produces the strongest downstream generalization.
Within the Qwen2.5-Math settings tested, RLVR data selection appears regime-dependent rather than reducible to a static ranking of problem quality.
\end{abstract}

% ============================================================
% §1 INTRODUCTION
% ============================================================

\section{Introduction}
\label{sec:intro}

If a language model's internal reasoning circuits can be identified, the most intuitive use for data selection is clear: train on problems that activate those circuits most strongly.
We test that intuition on the Qwen2.5-Math family and find that it does not hold.
On Qwen2.5-Math-7B, training on the problems that \emph{least} engage 46 heads identified by contrastive ablation improves over both random selection and the high-engagement alternative on three medium-difficulty benchmarks.
The result is not a simple reversal: the useful direction of selection depends on model capacity, source-pool diversity, and the target evaluation regime, in ways that a static ranking of problem quality does not capture.

Existing RLVR data selection methods rely on external criteria: difficulty filtering, hand-curation for ``hard but solvable'' problems \citep{limr2025}, reward-trajectory scoring, or single-example selection \citep{wang2025oneshot}.
What these approaches share is that they treat a problem's training value as an \emph{intrinsic property}---a function of the problem alone, independent of the model that will learn from it.
Yet the same problem can lie at the frontier of one model's competence and far beyond another's, and the same training prompt can produce informative gradients for one policy and zero gradients for another.
A selection signal that reflects the \emph{relationship} between a problem and the model's current reasoning state could provide more principled guidance.
We propose \emph{Circuit Reasoning Score} (CRS), a per-problem signal derived from the model's own reasoning circuitry.
Through contrastive ablation---comparing LM loss increases on reasoning versus trivial probes---we identify 46 reasoning-sensitive heads (bootstrap Jaccard = 0.91).
CRS measures how strongly these heads activate during a single forward pass, residualized against token length, requiring no reward labels, rollouts, or reference model.

The core empirical finding runs against the na\"ive high-activation hypothesis.
CRS-bottom selection improves over random on GSM8K, OlympiadBench, and Minerva Math by $+2.0$, $+1.6$, and $+2.9$\,pp; CRS-top's gain is smaller and not separable from CRS-middle.
This pattern has boundary conditions: it requires pool diversity (no detectable effect on a domain-curated Olympiads pool), the useful direction differs at 1.5B, and it decouples from training reward (the lowest-reward condition generalizes best).

Our contributions are:

(1) We propose CRS, a training-free, model-internal data selection signal computed from 46 reasoning-sensitive attention heads via a single forward pass on the frozen base model (\S\ref{sec:method}).

(2) Across 58 GRPO training runs on two model scales, we find evidence against the hypothesis that higher reasoning-circuit engagement produces better RLVR training data (\S\ref{sec:experiments}).

(3) We show that, within the settings tested, RLVR data selection is regime-dependent---modulated by pool diversity, model capacity, and evaluation difficulty---rather than reducible to a static ranking of problem quality (\S\ref{sec:experiments}--\ref{sec:discussion}).
% \end{enumerate}

% ============================================================
% §2 RELATED WORK (compressed, all citations preserved)
% ============================================================

\section{Related Work}
\label{sec:related}

\paragraph{RLVR for mathematical reasoning.}
RLVR builds on verifier-based reasoning and process supervision \citep{luo2024omegaprm,setlur2024rewardingprogress,yuan2024freeprocess,zhang2025lessonsprm}.
Methods such as GRPO \citep{grpo2024,deepseekr1} use within-group reward variance to train chain-of-thought reasoning, with improvements to stability \citep{dapo2025} and gradient bias \citep{drgrpo2025}.
A key property is that prompts whose completions all receive identical rewards provide no gradient signal, making prompt selection central.
\citet{yue2025doesrl} show RLVR narrows the output distribution; \citet{wang2025oneshot} show a single example can recover most gains.
Answer-level rewards can implicitly incentivize correct reasoning \citep{wen2025rlvr}, yet gains arise even under spurious rewards \citep{shao2025spurious}, suggesting performance depends on the model-data regime.

\paragraph{Data selection for RLVR.}
\citet{limr2025} select 1,389 of 8,523 Olympiad problems by reward-trajectory alignment; others use difficulty filtering \citep{rho12024} or dataset scaling \citep{dapo2025}; for SFT, \citet{limo2025} show 817 examples suffice.
Model-aware selection has been studied via influence-based methods \citep{xia2024less,engstrom2024dsdm,gu2024pds} and weak-to-strong filtering \citep{li2024superfiltering,mekala2024smallerselect}, but under supervised losses rather than RLVR.
Our work differs from curriculum learning \citep{bengio2009curriculum} and RLVR curricula that adjust sampling by reward variance or environment difficulty \citep{jiang2025vcrl,yang2025depthbreadth,zeng2025rlve}: CRS is not a difficulty proxy ($R^2 < 2\%$; \S\ref{sec:not_difficulty}), and its useful direction differs across the two scales tested.
Two RLVR-specific gaps remain: pool quality vs.\ selection quality are confounded without same-pool controls (\S\ref{sec:tier4}), and optimal selection may depend on model capacity (\S\ref{sec:cross_scale}).

\paragraph{Circuit-based data selection.}
Mechanistic interpretability has identified task-specific heads for induction \citep{olsson2022induction}, factual recall \citep{meng2022locating}, and reasoning \citep{elhage2021circuits}, with causal-intervention methods providing a broader toolkit \citep{hsu2024cdt,heimersheim2024activationpatching}.
Recent work examines head-level causal roles \citep{kissane2024attnsae,nam2025causalheadgating} and has uncovered interpretable circuits for chain-of-thought, syllogistic inference, and number representation \citep{dutta2024stepbystep,kim2025reasoningcircuits,kantamneni2025addition}.
\citet{circuitseer2025} applied circuit signals to SFT data selection via attention-pattern variance.
We depart in three ways: targeting RLVR rather than SFT; using contrastive ablation to isolate reasoning-specific heads; and testing both CRS directions rather than assuming higher engagement is better.

\paragraph{Position of this work.}
Prior RLVR selection asks \emph{which} data to choose; circuit-based selection asks \emph{whether} internal signals can rank examples (answered for SFT).
We ask \emph{when} a model-internal ranking helps, fails, or reverses---finding that the answer depends on pool diversity, model capacity, and evaluation regime.
This reframes RLVR data selection from a static ranking problem to a regime-dependent matching problem.

% ============================================================
% §3 METHOD
% ============================================================

\section{Method}
\label{sec:method}

\begin{figure*}[t]
  \centering
  \includegraphics[width=0.95\textwidth]{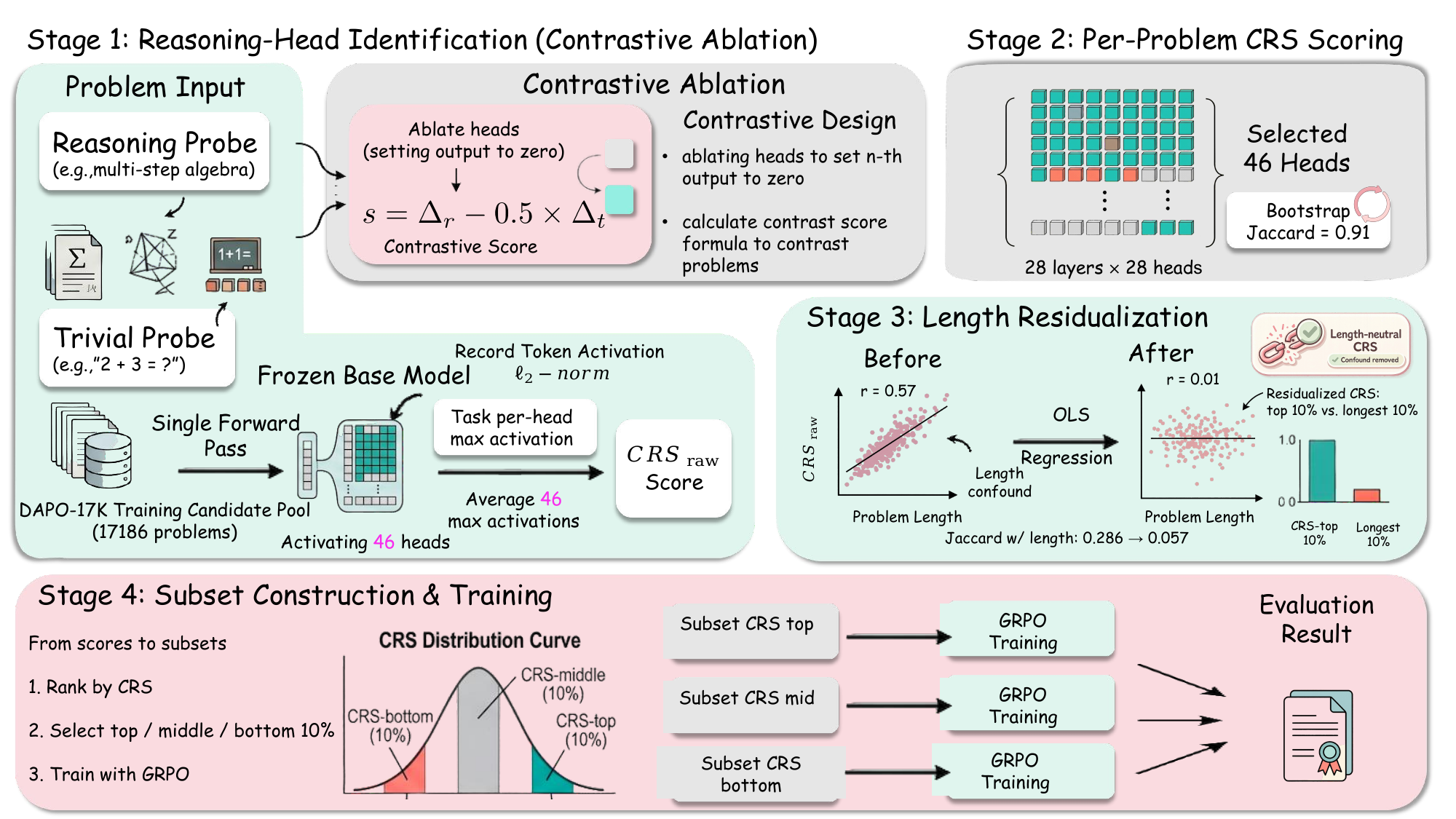}
  \caption{CRS pipeline.
  \textbf{Stage~1}: Contrastive ablation identifies 46 reasoning-sensitive heads (Jaccard = 0.91).
  \textbf{Stage~2}: Single forward pass records activation magnitudes; per-head max averaged to $\text{CRS}_{\text{raw}}$.
  \textbf{Stage~3}: OLS residualization removes length confound ($r: 0.57 \to 0.01$).
  \textbf{Stage~4}: Top/middle/bottom 10\% deciles form GRPO training subsets.}
  \label{fig:method}
\end{figure*}

Our starting point is a simple observation: if certain attention heads specialize in mathematical reasoning, then how strongly a problem engages those heads should reflect the problem's relationship to the model's reasoning capacity.
A problem that barely activates reasoning heads may lie outside the model's usual repertoire; one that strongly activates them may be well within it.
Both ends could carry useful information for RLVR, but extracting a clean signal requires solving three problems: distinguishing \emph{reasoning-specific} heads from generally important heads (\S\ref{sec:circuit_id}), aggregating head-level activations into a per-problem score (\S\ref{sec:crs_compute}), and removing a length confound that otherwise dominates the signal (\S\ref{sec:resid}).
The result is CRS: a real-valued per-problem score from a single forward pass on the frozen base model, with no reward labels or rollouts required.
Figure~\ref{fig:method} illustrates the pipeline.

\subsection{Reasoning-Head Identification}
\label{sec:circuit_id}

Qwen2.5-Math-7B \citep{yang2024qwen25math} has 784 attention heads ($28 \times 28$, $d_h{=}128$).
To isolate heads specifically important for reasoning (not just general language modeling), we use contrastive ablation.
We curate 50 reasoning probes (Olympiad-style) and 50 trivial probes (single-step arithmetic), held out from all training and evaluation data.
For each head, we zero its output and measure LM loss increases on each probe set ($\Delta_r$, $\Delta_t$):
\begin{equation}
\label{eq:contrastive}
s_{l,h} = \Delta_r(l,h) - 0.5 \, \Delta_t(l,h).
\end{equation}
The contrastive term is necessary because $\Delta_r$ and $\Delta_t$ correlate at $r \approx 0.75$; Eq.~\ref{eq:contrastive} extracts reasoning-specific importance beyond general-purpose function, unlike \citet{circuitseer2025} who ablate against reasoning probes only.

We repeat the scan three times with resampled probes.
The intersection yields 47 stable heads (pairwise Jaccard 0.89--0.92, mean 0.91; cross-round score correlation $r = 0.9998$).
Excluding one global-bottleneck outlier (L0H3, score $4.8\times$ the runner-up), the final set $\mathcal{H}_r$ contains \textbf{46 heads} spanning 21 of 28 layers.

We use ``reasoning-sensitive heads'' to denote heads identified by contrastive ablation on held-out probes; we do not claim a verified causal pathway.

\subsection{CRS Computation}
\label{sec:crs_compute}

We opt for activation magnitude---the $\ell_2$-norm of each head's output---rather than attention-weight statistics, since activation norms directly measure how much information a head contributes to the residual stream \citep{kissane2024attnsae,nam2025causalheadgating}.
For each problem $x$ with $n$ tokens, a single forward pass records each reasoning head's output activation at every position.
The raw score averages per-head maxima:
\begin{equation}
\label{eq:crs}
\text{CRS}_{\text{raw}}(x)
= \frac{1}{|\mathcal{H}_r|}
\sum_{h \in \mathcal{H}_r}
\max_{t \in [1,n]}
\bigl\| \mathbf{a}_h^{(t)}(x) \bigr\|_2.
\end{equation}
Taking the maximum captures peak reasoning engagement rather than diluting signal across boilerplate tokens.

\subsection{Length Residualization}
\label{sec:resid}

Raw CRS correlates with input length at $r = 0.568$.
Since length is a $2.8\times$ stronger difficulty predictor than CRS ($R^2 = 4.36\%$ vs.\ $1.57\%$; \S\ref{sec:not_difficulty}), we residualize via OLS on $\log n$:
\begin{equation}
\label{eq:resid}
\text{CRS}(x) = \text{CRS}_{\text{raw}}(x) - \hat\beta_0 - \hat\beta_1 \log n_x,
\end{equation}
yielding $r = 0.013$ with length.
The Jaccard overlap between the top-10\% by residualized CRS and the longest 10\% drops from 0.286 to 0.057, near chance.
All subsequent references to CRS denote this residualized version.
Appendix~\ref{app:sensitivity} reports sensitivity to the residualization form and to the contrastive weight in Eq.~\ref{eq:contrastive}.

\subsection{Selection and Controls}
\label{sec:selection}

We rank DAPO-Math-17K \citep{dapo2025} (17,186 problems) by CRS and extract \textbf{CRS-top} (highest 10\%), \textbf{CRS-middle} (percentiles 45--55), and \textbf{CRS-bottom} (lowest 10\%; 1,718 each), plus per-seed \textbf{Random} baselines.

\paragraph{Random-heads control.}
\label{sec:method:randheads}
We repeat scoring with 46 randomly sampled non-reasoning heads and train on the resulting top-10\% (\textbf{RandHeads}), testing whether the specific 46-head identity matters.

\paragraph{Cross-pool control.}
We re-score the 8,523-problem Olympiads pool---from which LIMR \citep{limr2025} was curated---and construct size-matched subsets (1,389 each).
Jaccard with LIMR's selection is 0.15--0.17, confirming independence.

\paragraph{Cross-scale control.}
We replicate on Qwen2.5-Math-1.5B, adding \texttt{xmod}: data selected by 7B's CRS, used to train 1.5B.
\texttt{xmod} is a scale-transfer variant of the weak-to-strong direction \citep{li2024superfiltering,mekala2024smallerselect}, not a full weak-to-strong filtering baseline.

% ============================================================
% §4 EXPERIMENTS
% ============================================================

\section{Experiments}
\label{sec:experiments}

\subsection{Setup}
\label{sec:setup}

All experiments use GRPO \citep{grpo2024} via TRL~v1.3.0 with the hyperparameters in Table~\ref{tab:hparams}, following defaults from \citet{deepseekr1} and \citet{dapo2025} on the Qwen2.5-Math family \citep{yang2024qwen25math}.
Binary reward: 1 if the \texttt{\textbackslash boxed\{\}} answer matches ground truth (SymPy), 0 otherwise.
Under TRL's \texttt{RepeatSampler}, 1,000 steps cover ${\sim}$58\% of each 1,718-prompt subset.

\begin{table}[t]
  \centering\small
  \begin{tabular}{@{}ll@{}}
    \toprule
    \textbf{Parameter} & \textbf{Value} \\
    \midrule
    Learning rate & $1 \times 10^{-6}$, cosine, warmup 0.1 \\
    Max gradient norm & 1.0 \\
    Temperature / Generations & 0.7 / 16 per prompt \\
    Max completion length & 4,096 tokens \\
    Batch / Grad.\ accum. & 4 / 4 (eff.\ 16 prompts/step) \\
    Training steps & 1,000 \\
    Hardware & 8$\times$A100-SXM4-80GB \\
    \bottomrule
  \end{tabular}
  \caption{GRPO hyperparameters (shared across all conditions and scales).}
  \label{tab:hparams}
\end{table}

\paragraph{Conditions.}
\textbf{Tier~1} (5 seeds): CRS top / mid / bottom-10\%, Random 10\% of DAPO-17K.
\textbf{Tier~2} (3 seeds): LIMR \citep{limr2025} (1,389 from the 8,523-problem Olympiads pool), Full data (17K), RandHeads top-10\%, and a difficulty-filtering baseline (Diff-hard, 3 seeds; Diff-easy, 2 seeds; \S\ref{sec:baselines}).
\textbf{Tier~4} (3 seeds): Oly-CRS-top / bottom / Random (1,389 each from the Olympiads pool).
\textbf{1.5B} (3 seeds): CRS top / mid / bottom-10\%, Random 10\%, \texttt{xmod}.
Total: 58 runs.

\paragraph{Evaluation.}
Five primary benchmarks spanning a difficulty gradient (Table~\ref{tab:benchmarks}); OlympiadBench \citep{he2024olympiadbench} anchors the competition tier.
Two additional benchmarks (AMC~2023, $n{=}40$; AIME~2025, $n{=}30$) are in Appendix~\ref{app:full_table}.
Benchmarks with $n \leq 60$ use avg@8 ($\tau{=}0.6$); others use greedy pass@1.
Pairwise comparisons are reported at the seed unit (\S\ref{sec:bottom_wins}).

\begin{table}[t]
  \centering\small
  \begin{tabular}{@{}llcl@{}}
    \toprule
    \textbf{Benchmark} & \textbf{Difficulty} & $N$ & \textbf{Metric} \\
    \midrule
    GSM8K & Grade-school & 1,319 & pass@1 \\
    Minerva Math & Undergrad STEM & 272 & pass@1 \\
    MATH-500 & Competition (mixed) & 500 & pass@1 \\
    OlympiadBench & Competition (mixed) & 562 & pass@1 \\
    AIME 2024 & Competition (hard) & 60 & avg@8 \\
    \bottomrule
  \end{tabular}
  \caption{Primary benchmarks (no overlap with training pool).}
  \label{tab:benchmarks}
\end{table}

\subsection{CRS Is Not a Monotonic Quality Score}
\label{sec:main_results}

If CRS captured a monotonic notion of data quality, higher-CRS data should produce uniformly better models.
Table~\ref{tab:main} shows this is not the case.

\begin{table*}[t]
  \centering\small
  \caption{7B main results: 7 conditions $\times$ 5 benchmarks (mean $\pm$ std).
  Tier~1: $n{=}5$; Tier~2: $n{=}3$.
  \textbf{Bold}: highest Tier~1 mean.
  LIMR uses a different pool; see \S\ref{sec:tier4}.}
  \label{tab:main}
  \setlength{\tabcolsep}{5pt}
  \begin{tabular}{@{}llccccc@{}}
    \toprule
    & \textbf{Condition}
      & \textbf{GSM8K} & \textbf{Minerva} & \textbf{MATH-500} & \textbf{Olymp.}
      & \textbf{AIME\,'24} \\
    \midrule
    & Base & 81.20 & 16.18 & 66.80 & 37.76 & 11.46 \\
    \midrule
    \multirow{4}{*}{\rotatebox[origin=c]{90}{\scriptsize T1}}
      & CRS-top    & $84.26 \pm 0.64$ & $20.81 \pm 1.90$ & $69.20 \pm 1.51$ & $37.45 \pm 0.47$ & $16.63 \pm 1.79$ \\
      & CRS-mid    & $83.52 \pm 0.57$ & $20.88 \pm 2.13$ & $68.52 \pm 1.51$ & $37.87 \pm 0.88$ & $17.08 \pm 1.97$ \\
      & CRS-bot    & $\mathbf{85.81 \pm 2.48}$ & $\mathbf{22.50 \pm 3.05}$ & $68.56 \pm 3.88$ & $\mathbf{38.36 \pm 1.91}$ & $14.88 \pm 1.97$ \\
      & Random     & $83.82 \pm 0.58$ & $19.63 \pm 1.09$ & $69.12 \pm 1.29$ & $36.75 \pm 0.86$ & $16.50 \pm 1.26$ \\
    \midrule
    \multirow{3}{*}{\rotatebox[origin=c]{90}{\scriptsize T2}}
      & LIMR       & $85.39 \pm 1.03$ & $21.20 \pm 2.61$ & $70.93 \pm 0.64$ & $38.00 \pm 0.66$ & $15.41 \pm 0.91$ \\
      & Full (17K) & $83.73 \pm 0.61$ & $17.53 \pm 2.76$ & $67.93 \pm 0.70$ & $36.30 \pm 0.86$ & $17.50 \pm 1.46$ \\
      & RandHeads  & $85.65 \pm 2.72$ & $20.47 \pm 2.97$ & $69.67 \pm 2.72$ & $37.18 \pm 2.33$ & $15.76 \pm 2.65$ \\
    \bottomrule
  \end{tabular}
\end{table*}

\begin{figure}[t]
  \centering
  \includegraphics[width=\columnwidth]{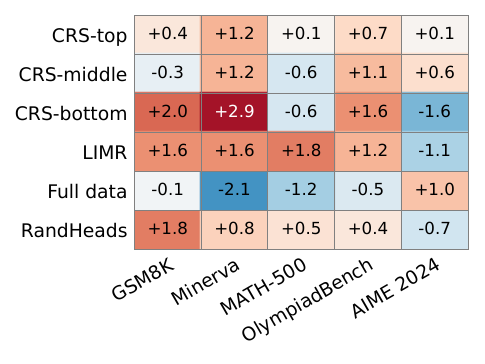}
  \caption{Gain over Random (pp) per condition per benchmark, as unpaired differences of condition means.
  CRS-bottom's gains cluster on medium-difficulty benchmarks; LIMR shows a deficit on AIME.}
  \label{fig:gain_heatmap}
\end{figure}

CRS-bottom achieves the highest Tier~1 accuracy on GSM8K, OlympiadBench, and Minerva, while CRS-top's margin over Random is small on every benchmark.
Figure~\ref{fig:gain_heatmap} summarizes this: CRS-bottom's gains (green) concentrate on medium-difficulty benchmarks; its deficit (red) appears on AIME.
No condition dominates all benchmarks.

\subsection{CRS-Bottom Improves Medium-Difficulty Reasoning}
\label{sec:bottom_wins}

We compare conditions at the seed unit, since what matters for a selection rule is whether its advantage reproduces across training runs.
CRS-bottom's advantage over Random is $+2.0$\,pp on GSM8K, $+2.9$\,pp on Minerva, and $+1.6$\,pp on OlympiadBench (Table~\ref{tab:seed_effect}), positive on 13 of 15 seed-by-benchmark cells and on all five seed composites, for a composite of $+2.15$\,pp.
A hierarchical binomial model of the same outcomes, treating the condition-by-seed interaction as a free parameter, places $91.3\%$ of the posterior for the condition effect above zero (Appendix~\ref{app:glmm}).
An item-level McNemar test on GSM8K ($p = 3.3\times10^{-4}$; Appendix~\ref{app:mcnemar}) confirms consistency across items within a run.

\begin{table}[t]
  \centering\small
  \setlength{\tabcolsep}{3pt}
  \caption{Seed-level paired differences against Random (pp, $n{=}5$ seeds).
  ``Pos.'': seed-by-benchmark cells favouring the condition, out of 15, over the three target benchmarks.}
  \label{tab:seed_effect}
  \resizebox{\linewidth}{!}{%
  \begin{tabular}{@{}lcccccc@{}}
    \toprule
    & \textbf{GSM8K} & \textbf{Minerva} & \textbf{MATH} & \textbf{Olymp.} & \textbf{AIME} & \textbf{Pos.} \\
    \midrule
    CRS-bot & $+2.0$ & $+2.9$ & $-0.6$ & $+1.6$ & $-1.6$ & 13\,/\,15 \\
    CRS-top & $+0.4$ & $+1.2$ & $+0.1$ & $+0.7$ & $+0.1$ & 10\,/\,15 \\
    CRS-mid & $-0.3$ & $+1.3$ & $-0.6$ & $+1.1$ & $+0.6$ & \phantom{0}8\,/\,15 \\
    \bottomrule
  \end{tabular}
  }
\end{table}

\begin{figure}[t]
  \centering
  \includegraphics[width=\columnwidth]{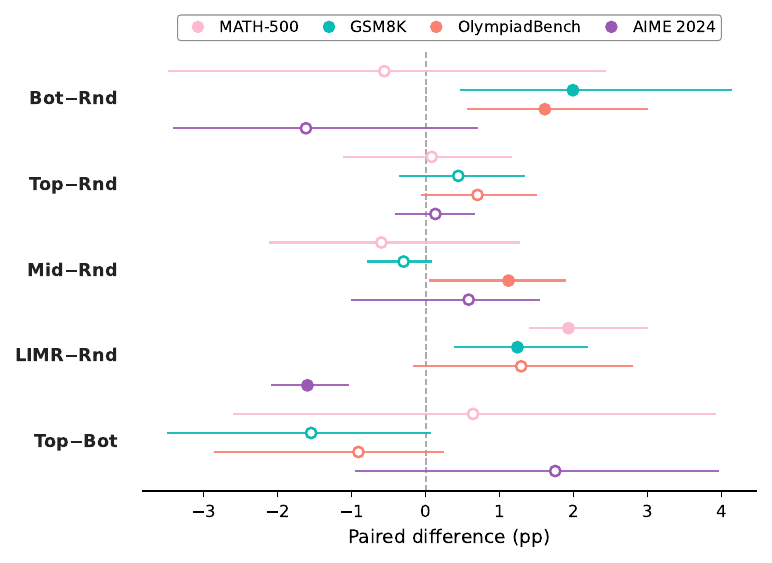}
  \caption{Seed-level paired differences for key comparisons across five benchmarks, with resampling envelopes.
  CRS-bottom exceeds Random on GSM8K, OlympiadBench, and Minerva; LIMR trails Random on AIME.}
  \label{fig:forest}
\end{figure}

The effect has clear boundaries: on MATH-500 no Tier~1 condition separates from Random, consistent with a ceiling effect at 66.8\% base accuracy; on AIME, CRS-bottom is $-1.6$\,pp against Random.
CRS-top's point estimate also exceeds Random on the three target benchmarks, so the relationship is not monotone in CRS, though the two upper conditions are not separable at this seed count (Appendix~\ref{app:ushape}).
The CRS-bottom advantage is specific to medium-difficulty evaluation---a boundary condition we revisit in \S\ref{sec:cross_scale}.
Figure~\ref{fig:forest} provides a visual summary of all key comparisons.

These results raise a natural question: is the CRS-bottom effect genuinely attributable to circuit-based selection, or could simpler explanations account for it?

\subsection{Ruling Out Alternative Explanations}
\label{sec:baselines}

We consider four alternative hypotheses, each addressed by a dedicated baseline.

\paragraph{``Reward-trajectory selection is all you need.''}
LIMR \citep{limr2025} selects by reward-trajectory alignment and serves as our reward-trajectory baseline.
It leads on MATH-500 ($+1.9$\,pp over Random) but trails Random on AIME ($-1.6$\,pp), where CRS-top exceeds it by $+1.7$\,pp.
LIMR's pool differs from ours (the 8,523-problem Olympiads pool vs.\ DAPO-17K), confounding selection quality with pool quality---a confound we isolate in \S\ref{sec:tier4}.
Selection criteria optimized for intermediate difficulty do not transfer to the hardest evaluation tier.

\paragraph{``Any head subset works.''}
RandHeads achieves similar means to CRS-top (GSM8K: 85.65 vs.\ 84.26) but with $2$--$4\times$ higher cross-seed variance (GSM8K std: 2.72 vs.\ 0.64; MATH: 2.72 vs.\ 1.51).
Arbitrary head subsets can produce effective selections by chance in some seeds, but the 46 identified heads provide a more \emph{consistent} signal.
A post-training ablation against six layer-matched random head sets (Appendix~\ref{app:ablation}) approaches the same question causally.

\paragraph{``More data is better.''}
Full-data training (17K problems, 5.8\% coverage at 1,000 steps) trails all 10\% subsets on MATH-500 ($67.93 \pm 0.70$).
A larger pool does not help when training budget is fixed; focused selection outperforms diluted sampling.

\paragraph{``Higher CRS is better.''}
CRS-top's margin over Random is small on every benchmark and not separable from CRS-middle (\S\ref{sec:bottom_wins}; full difference table in Appendix~\ref{app:full_table}).
The larger and more seed-stable effect at 7B lies in the bottom decile, not the top---a finding that runs against the na\"ive expectation that problems engaging the most reasoning circuitry should be the best training data.

\paragraph{``Difficulty filtering is all you need.''}
Using the rollout-estimated success rates of \S\ref{sec:not_difficulty} ($K{=}4$; valid estimates for 16,478 pool problems), we built two disjoint 1,647-problem subsets at the two ends of the SR distribution---\textbf{Diff-hard} and \textbf{Diff-easy}, selected by rank and labelled by measured mean SR---and trained them under the identical protocol (Appendix~\ref{app:difficulty}).
CRS-bottom remains the strongest condition on GSM8K, Minerva, and OlympiadBench (Table~\ref{tab:difficulty}), the three benchmarks on which our claim rests.
The two selectors are nearly orthogonal (Jaccard $0.030$--$0.069$ against a chance level of ${\approx}0.051$), independently reproducing the result of \S\ref{sec:not_difficulty} that CRS is not a difficulty proxy.
The cost asymmetry is large: difficulty filtering required 65,912 generations, against a single forward pass per problem for CRS.

\begin{table}[t]
  \centering\small
  \setlength{\tabcolsep}{3pt}
  \caption{Difficulty-filtering baseline (mean $\pm$ std).
  Conditions labelled by measured mean success rate.
  CRS-bottom and Random rows are the Tier~1 references.}
  \label{tab:difficulty}
  \resizebox{\linewidth}{!}{%
  \begin{tabular}{@{}lcccccc@{}}
    \toprule
    \textbf{Condition} & $n$ & \textbf{GSM8K} & \textbf{Minerva} & \textbf{MATH-500} & \textbf{Olymp.} & \textbf{AIME\,'24} \\
    \midrule
    Diff-hard (SR${=}$.25) & 3 & $84.53{\pm}0.69$ & $21.08{\pm}1.50$ & $68.60{\pm}1.10$ & $37.47{\pm}0.90$ & $17.57{\pm}2.10$ \\
    Diff-easy (SR${=}$.92) & 2 & $83.36{\pm}0.48$ & $18.75{\pm}0.00$ & $68.10{\pm}0.14$ & $36.98{\pm}0.37$ & $15.94{\pm}1.00$ \\
    \midrule
    CRS-bottom & 5 & $85.81{\pm}2.48$ & $22.50{\pm}3.05$ & $68.56{\pm}3.88$ & $38.36{\pm}1.91$ & $14.88{\pm}1.97$ \\
    Random & 5 & $83.82{\pm}0.58$ & $19.63{\pm}1.09$ & $69.12{\pm}1.29$ & $36.75{\pm}0.86$ & $16.50{\pm}1.26$ \\
    \bottomrule
  \end{tabular}
  }
\end{table}

Taken together, these baselines indicate that CRS-bottom's advantage is not reducible to reward-trajectory selection, arbitrary head signals, data volume, CRS magnitude, or difficulty filtering.
But a deeper question remains: is the effect a property of CRS itself, or of the training pool?

\subsection{Selection Effects Require Pool Diversity}
\label{sec:tier4}

On the homogeneous 8,523-problem Olympiads pool, we construct three size-matched subsets (1,389 each).
Table~\ref{tab:tier4} shows no detectable difference among the three, despite selecting nearly independent problem sets (Jaccard with LIMR: 0.15--0.17).

\begin{table}[t]
  \centering\small
  \caption{Tier~4: Olympiads pool ($n{=}3$ seeds).}
  \label{tab:tier4}
  \begin{tabular}{@{}lccc@{}}
    \toprule
    & \textbf{MATH} & \textbf{GSM8K} & \textbf{Olymp.} \\
    \midrule
    Oly-CRS-top & $70.5 \pm 2.5$ & $86.0 \pm 2.7$ & $37.9 \pm 0.8$ \\
    Oly-CRS-bot & $71.4 \pm 3.1$ & $86.6 \pm 2.9$ & $38.1 \pm 3.3$ \\
    Oly-Random  & $69.7 \pm 1.4$ & $86.8 \pm 1.0$ & $38.1 \pm 1.6$ \\
    \midrule
    LIMR (ref.) & $70.9 \pm 0.6$ & $85.4 \pm 1.0$ & $38.0 \pm 0.7$ \\
    \bottomrule
  \end{tabular}
\end{table}

On diverse DAPO-17K, CRS stratification produces up to 2.9\,pp differences; on the narrow Olympiads pool, it produces none.
The contrast is informative: on the diverse pool, CRS stratification separates grade-school arithmetic from Olympiad reasoning, and training outcomes diverge; on the homogeneous pool, CRS stratification produces subsets that differ in identity but converge in outcome.
This is consistent with adaptive-environment work showing that narrow distributions lose learning signal \citep{zeng2025rlve,stojanovski2025reasoninggym}.
Pool diversity is a prerequisite for selection effects---if the pool is already domain-curated, effort is better spent expanding it than optimizing within it.

This establishes that CRS effects are pool-dependent.
But are they also model-dependent?

\subsection{The Useful Direction Differs at 1.5B}
\label{sec:cross_scale}

\begin{figure}[t]
  \centering
  \includegraphics[width=\columnwidth]{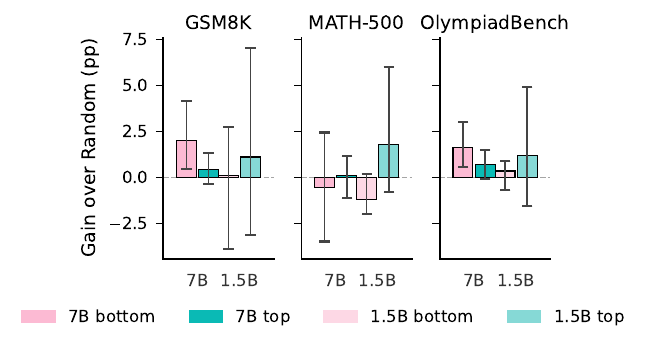}
  \caption{Useful direction at the two scales tested.
  At 7B, CRS-bottom (pink) is positive; at 1.5B, CRS-top (teal) leads.
  Error bars: resampling envelopes.}
  \label{fig:scale_direction}
\end{figure}

At 1.5B (Table~\ref{tab:1_5b}), CRS-top leads on MATH-500 (65.4), GSM8K (77.5), and OlympiadBench (31.4)---the same benchmarks where CRS-bottom led at 7B.
CRS-bottom trails CRS-top by $3.0$\,pp on MATH-500, against $-0.6$\,pp at 7B.
Figure~\ref{fig:scale_direction} visualizes the difference.

\begin{table}[t]
  \centering\small
  \caption{1.5B results ($n{=}3$ seeds).
  \textbf{Bold}: best per benchmark.
  \texttt{xmod}: 7B's CRS top-10\% used for 1.5B.}
  \label{tab:1_5b}
  \setlength{\tabcolsep}{3.5pt}
  \begin{tabular}{@{}lcccc@{}}
    \toprule
    & \textbf{MATH} & \textbf{GSM8K} & \textbf{Olymp.} & \textbf{AIME} \\
    \midrule
    Base & 63.20 & 76.19 & 30.94 & 7.50 \\
    \midrule
    CRS-top    & $\mathbf{65.4{\pm}3.0}$ & $\mathbf{77.5{\pm}4.5}$ & $\mathbf{31.4{\pm}3.3}$ & $9.9{\pm}1.3$ \\
    CRS-bot    & $62.4{\pm}0.6$ & $76.5{\pm}3.1$ & $30.6{\pm}0.5$ & $10.2{\pm}1.3$ \\
    Random     & $63.6{\pm}0.8$ & $76.4{\pm}0.9$ & $30.2{\pm}0.5$ & $9.5{\pm}0.5$ \\
    \texttt{xmod} & $63.1{\pm}1.2$ & $77.3{\pm}0.9$ & $30.3{\pm}0.2$ & $9.7{\pm}1.2$ \\
    \bottomrule
  \end{tabular}
\end{table}

The reversal is consistent with GRPO's reward-variance mechanism \citep{jiang2025vcrl,yang2025depthbreadth}: GRPO produces nonzero gradients only when the $G{=}16$ completions exhibit reward variance---some succeed and others fail.
At 7B, both CRS-top and CRS-bottom problems lie within the regime of useful reward variance, but CRS-bottom provides more novel reasoning patterns when successes occur, producing transferable skills for easier benchmarks.
At 1.5B, CRS-bottom problems fall below the threshold where the weaker model can generate correct completions, yielding predominantly zero-advantage steps.
CRS-top problems remain within the 1.5B model's productive zone, explaining why high-CRS data becomes more valuable at smaller scale.
This interpretation predicts that the optimal CRS direction should shift continuously with model capacity---a prediction we leave to future verification with intermediate-scale models.

\paragraph{Cross-model portability.}
The \texttt{xmod} condition matches native 1.5B CRS-top (GSM8K: 77.3 vs.\ 77.5, std 0.87 vs.\ 4.48) and trends positive over Random ($+1.0$\,pp), echoing weak-to-strong filtering results \citep{li2024superfiltering,mekala2024smallerselect} and suggesting CRS captures a partially transferable pool-level signal.
CRS subsets computed once on a strong base model can be reused across moderate scale shifts.

\subsection{Training Reward Does Not Predict Generalization}
\label{sec:dynamics}

\begin{figure}[t]
  \centering
  \includegraphics[width=\columnwidth]{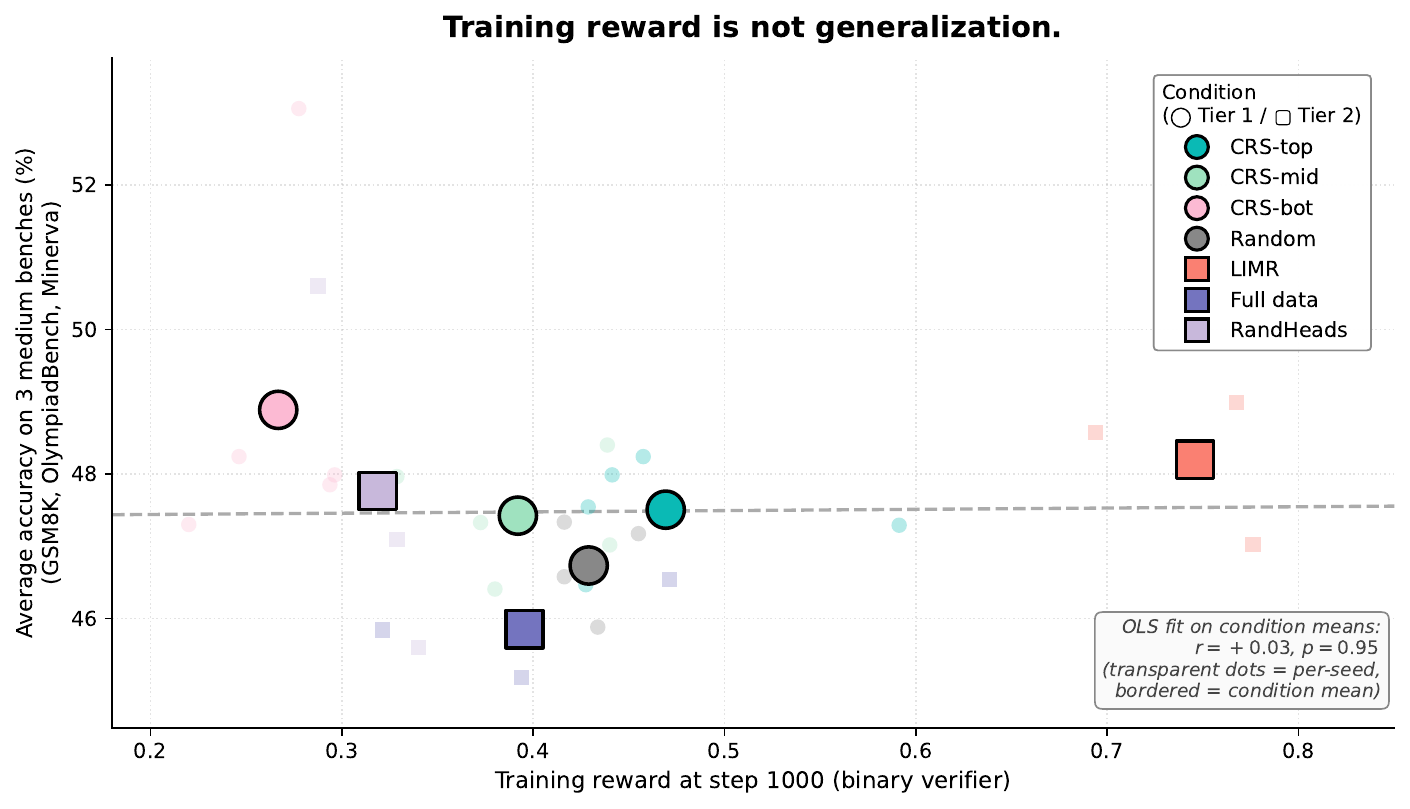}
  \caption{Training reward vs.\ generalization.
  Across all runs $r = -0.18$; across the seven condition means $r = +0.03$.
  Optimizing for training reward does not predict downstream capability.}
  \label{fig:reward_gen}
\end{figure}

If training reward were a sufficient proxy for useful learning, the highest-reward condition should generalize best.
The opposite emerges: CRS-bottom converges to the lowest reward ($0.265$) yet achieves the highest GSM8K ($85.81$) and Minerva ($22.50$); LIMR reaches the highest reward ($0.775$) but underperforms Random on AIME (Table~\ref{tab:dynamics}).
Across all runs the correlation between training reward and downstream accuracy is $r = -0.18$; across condition means it is $+0.03$ (Figure~\ref{fig:reward_gen}).

\begin{table}[t]
  \centering\small
  \caption{Training dynamics at step 1,000 (7B, seed-averaged).}
  \label{tab:dynamics}
  \begin{tabular}{@{}lcc@{}}
    \toprule
    \textbf{Condition} & \textbf{Reward} & \textbf{Zero-std \%} \\
    \midrule
    LIMR & 0.775 & $\sim$7\% \\
    CRS-middle & 0.485 & $\sim$24\% \\
    Random & 0.470 & $\sim$40\% \\
    CRS-top & 0.362 & $\sim$44\% \\
    RandHeads & 0.296 & $\sim$33\% \\
    CRS-bottom & \textbf{0.265} & $\sim$28\% \\
    Full data & 0.200 & $\sim$53\% \\
    \bottomrule
  \end{tabular}
\end{table}

This decoupling resolves an apparent paradox: how can CRS-bottom---a condition where the model mostly \emph{fails} during training---produce the strongest downstream performance?
Training reward measures how well the model solves its training problems; it does not measure whether the learned skills generalize.
CRS-bottom forces exploration of reasoning patterns outside the model's existing repertoire, and the resulting skills transfer effectively to evaluation benchmarks where those patterns are useful.
Practitioners should not optimize data selection for training reward: in RLVR, the conditions that look worst during training may generalize best.

% ============================================================
% §5 DISCUSSION
% ============================================================

\begin{figure*}[t]
  \centering
  \includegraphics[width=0.9\textwidth]{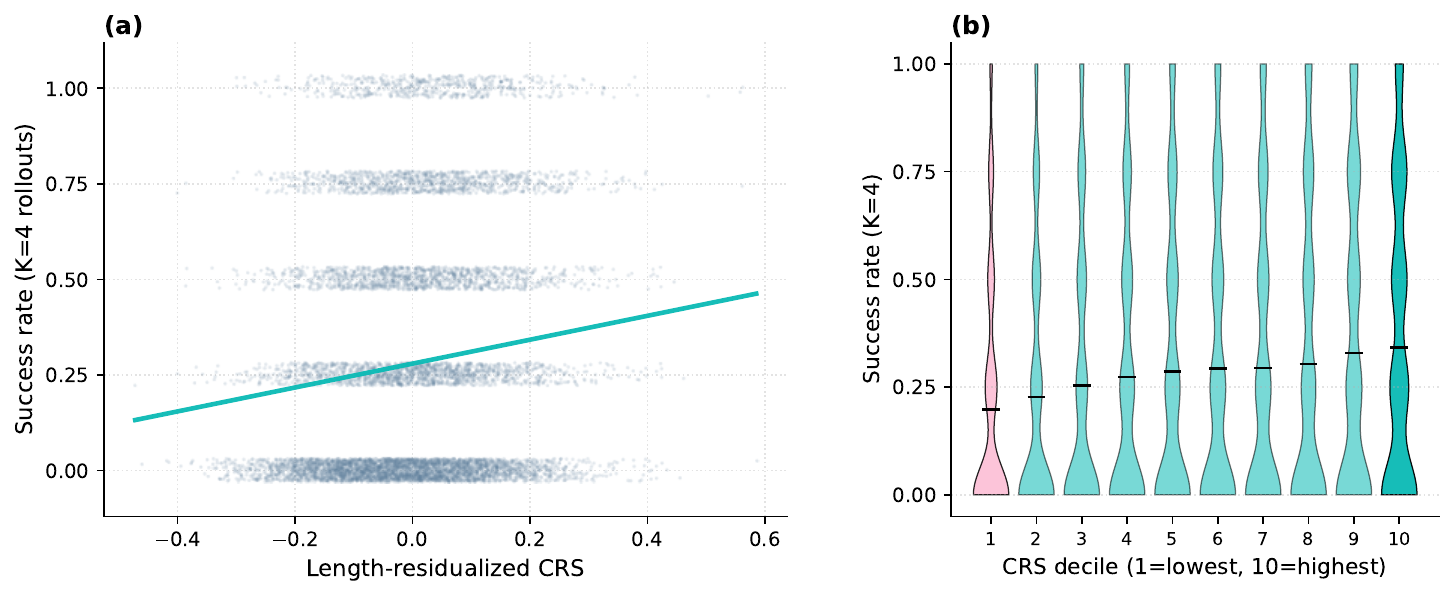}
  \caption{CRS is not a difficulty proxy.
  \textbf{(a)}~Scatter of length-residualized CRS vs.\ rollout-estimated success rate ($K{=}4$, $n{=}16{,}478$).
  The regression line (Spearman $\rho{=}0.125$, $R^2{=}1.57\%$) is barely visible against the spread.
  \textbf{(b)}~Violin plots of SR per CRS decile.
  Distributions overlap heavily; median success rate shifts only from ${\sim}0.15$ to ${\sim}0.30$ across deciles.
  CRS and difficulty rank problems in nearly independent orders.}
  \label{fig:crs_vs_sr}
\end{figure*}

\section{Discussion}
\label{sec:discussion}

\subsection{CRS Is Not a Difficulty Proxy}
\label{sec:not_difficulty}

A natural concern is that CRS simply recapitulates problem difficulty: if low-CRS problems are easy and high-CRS problems are hard, the Tier~1 effects could reduce to ``train on easy problems to improve easy benchmarks.''
We test this directly by computing rollout-estimated success rates (SR) for 16,478 DAPO-17K problems ($K{=}4$ rollouts at $\tau{=}0.6$) and measuring their correlation with residualized CRS.

\begin{table}[t]
  \centering\small
  \caption{Correlation with base-model success rate ($n{=}16{,}478$).}
  \label{tab:rho}
  \begin{tabular}{@{}lrr@{}}
    \toprule
    \textbf{Signal} & \textbf{$\rho$} & \textbf{$R^2$ (\%)} \\
    \midrule
    CRS (residualized) & $+0.125$ & 1.57 \\
    $\log$ token length & $-0.209$ & 4.36 \\
    \bottomrule
  \end{tabular}
\end{table}

Residualized CRS explains only 1.57\% of the variance in base-model success rate ($\rho = 0.125$, 95\% CI $[+0.112, +0.140]$, $n{=}16{,}478$); token length explains $4.36\%$---a $2.8\times$ stronger predictor (Table~\ref{tab:rho}).
The difficulty-filtering baseline of \S\ref{sec:baselines} reproduces this independently: its SR-selected subsets overlap the CRS subsets at Jaccard $0.030$--$0.069$, at chance.
Figure~\ref{fig:crs_vs_sr} makes the weak relationship visually apparent.
The scatter (panel a) shows 16,478 problems as a diffuse cloud with a barely perceptible positive slope; SR takes all five discrete values at every CRS level.
The per-decile violin plot (panel b) confirms this: the SR distribution within each CRS decile spans the full $[0, 1]$ range, with median SR shifting only from ${\sim}0.15$ (decile~1) to ${\sim}0.30$ (decile~10).
As a further check, Jaccard overlap between CRS-top and SR-top (the easiest 10\%) is 0.078, barely above the chance level of 0.053; CRS-bottom and SR-bottom overlap at 0.055---essentially random.
CRS-bottom's advantage on medium-difficulty benchmarks (\S\ref{sec:bottom_wins}) therefore cannot be attributed to ``CRS selects easy problems'': CRS-bottom and the easiest-by-SR problems are almost entirely disjoint sets.

\subsection{Three Axes of Selection Sensitivity}
\label{sec:three_axes}

The experiments in \S\ref{sec:experiments} reveal that RLVR data selection effects are modulated by three factors, each of which can independently determine whether selection helps, hurts, or is irrelevant.

\paragraph{Pool diversity is the prerequisite.}
On diverse DAPO-17K, CRS-bottom outperforms Random by up to 2.9\%; on the domain-narrow Olympiads pool, the same procedure produces no detectable effect (\S\ref{sec:tier4}).
If the training pool has already been curated to a narrow domain, effort is better spent expanding diversity than optimizing within-pool selection.

\paragraph{Evaluation regime determines the useful direction.}
CRS-bottom's advantage concentrates on medium-difficulty benchmarks (GSM8K, OlympiadBench, Minerva); on AIME, it provides no benefit.
No single direction dominates all evaluation regimes---the practical guidance is to match selection direction to the target.

\paragraph{Model capacity shifts the productive zone.}
At 7B, the model benefits from low-CRS problems---unfamiliar territory that forces exploration.
At 1.5B, the same problems fall below the productive reward-variance threshold (\S\ref{sec:cross_scale}), and high-CRS problems become more valuable.

\paragraph{Unified framing.}
These three axes reframe the question from ``which selection method is best?''\ to ``which regime am I in?''\ A diverse pool with a capable model and medium-difficulty targets is the regime where CRS-bottom is most effective.
Outside this regime, the signal either vanishes or reverses, and practitioners should adjust accordingly.

% ============================================================
% §6 CONCLUSION
% ============================================================

\section{Conclusion}
\label{sec:conclusion}

Within the Qwen2.5-Math settings tested, the training value of a problem for RLVR does not appear to be an intrinsic property of the problem.
CRS serves less as a universal selector and more as a probe that exposes how selection effects depend on context---appearing on a diverse pool but not detectable on a curated one, favouring low-engagement data at 7B but high-engagement data at 1.5B, and producing the strongest generalization from the lowest-reward condition.
These interactions are invisible to criteria that treat data quality as problem-intrinsic, which is why selection optimized for intermediate difficulty can underperform random selection on the hardest tier.
For practitioners working in a comparable setting: on a diverse pool with a 7B-scale model, CRS-bottom improved medium-difficulty reasoning in our runs; on curated pools or smaller models, the signal was not detectable or ran the other way.

% ============================================================
% LIMITATIONS (standalone, before references, does NOT count toward 8 pages)
% ============================================================

\section*{Limitations}

\paragraph{Scope.}
All experiments use the Qwen2.5-Math family at 7B and 1.5B, on mathematical data only.
Generalization to other architectures and verifiable-reward domains is untested.

\paragraph{Statistical power and prompt coverage.}
The 7B analysis uses $n{=}5$ seeds and Tier 4 / 1.5B use $n{=}3$; we report effect sizes rather than significance labels.
At 1,000 steps each run sees 58.2\% of its 1,718-prompt subset, so labels denote a random sample of each decile, contributing to cross-seed variance (Appendix~\ref{app:glmm}).

\paragraph{Causal interpretation.}
CRS is correlational: activation magnitudes at heads identified via ablation.
Zero-ablating these heads removes the CRS-bottom advantage but is equally destructive to the untrained base model, indicating a pathway already present in the base model rather than one built by GRPO.
We present CRS as a selection-level predictive signal rather than a verified reasoning circuit.

\section*{Ethics Statement}
This work uses publicly available mathematical benchmarks and open-weight models.
No human subjects, private data, or dual-use concerns are involved.

\section*{Acknowledgments}
During the preparation of this work, the authors used AI-assisted tools for language polishing, grammar checking, coding assistance, and brainstorming.
All research ideas, experimental designs, data analyses, and conclusions were developed and verified by the authors.
The authors take full responsibility for the content of the paper.

This work was supported in part by the National Natural Science Foundation of China under Grant 62477001.

\bibliography{custom}
% ============================================================
% APPENDIX
% ============================================================

\appendix

% ============================
\section{Full Results Tables}
\label{app:full_table}

\subsection{7B Main Grid: All 7 Benchmarks}
\label{app:full_7bench}

Table~\ref{tab:full_7bench} extends Table~\ref{tab:main} with two additional benchmarks: AMC~2023 ($n{=}40$, avg@8) and AIME~2025 ($n{=}30$, avg@8).
These benchmarks have high per-problem variance due to small $n$ and are excluded from the primary analysis in the main text.

\begin{table*}[!htbp]
  \centering\small
  \caption{Full 7B results: 7 conditions $\times$ 7 benchmarks (mean $\pm$ std across seeds).
  Tier~1: $n{=}5$ seeds; Tier~2: $n{=}3$.
  AMC~2023 and AIME~2025 (rightmost two columns) are excluded from primary analysis due to small $n$.}
  \label{tab:full_7bench}
  \setlength{\tabcolsep}{3.5pt}
  \begin{tabular}{@{}llccccccc@{}}
    \toprule
    & \textbf{Condition}
      & \textbf{GSM8K} & \textbf{Minerva} & \textbf{MATH-500} & \textbf{Olymp.}
      & \textbf{AIME\,'24} & \textbf{AMC\,'23} & \textbf{AIME\,'25} \\
    \midrule
    & Base & 81.20 & 16.18 & 66.80 & 37.76 & 11.46 & 50.31 & 7.92 \\
    \midrule
    \multirow{4}{*}{\rotatebox[origin=c]{90}{\scriptsize Tier 1}}
      & CRS-top     & $84.26{\pm}0.64$ & $20.81{\pm}1.90$ & $69.20{\pm}1.51$ & $37.45{\pm}0.47$ & $16.63{\pm}1.79$ & $60.19{\pm}2.01$ & $9.92{\pm}1.65$ \\
      & CRS-mid     & $83.52{\pm}0.57$ & $20.88{\pm}2.13$ & $68.52{\pm}1.51$ & $37.87{\pm}0.88$ & $17.08{\pm}1.97$ & $61.56{\pm}3.17$ & $9.75{\pm}2.14$ \\
      & CRS-bot     & $85.81{\pm}2.48$ & $22.50{\pm}3.05$ & $68.56{\pm}3.88$ & $38.36{\pm}1.91$ & $14.88{\pm}1.97$ & $59.50{\pm}1.37$ & $9.58{\pm}2.69$ \\
      & Random      & $83.82{\pm}0.58$ & $19.63{\pm}1.09$ & $69.12{\pm}1.29$ & $36.75{\pm}0.86$ & $16.50{\pm}1.26$ & $58.81{\pm}1.42$ & $9.08{\pm}1.00$ \\
    \midrule
    \multirow{3}{*}{\rotatebox[origin=c]{90}{\scriptsize Tier 2}}
      & LIMR        & $85.39{\pm}1.03$ & $21.20{\pm}2.61$ & $70.93{\pm}0.64$ & $38.00{\pm}0.66$ & $15.41{\pm}0.91$ & $58.54{\pm}3.21$ & $10.56{\pm}1.46$ \\
      & Full (17K)  & $83.73{\pm}0.61$ & $17.53{\pm}2.76$ & $67.93{\pm}0.70$ & $36.30{\pm}0.86$ & $17.50{\pm}1.46$ & $59.27{\pm}1.09$ & $10.00{\pm}0.42$ \\
      & RandHeads   & $85.65{\pm}2.72$ & $20.47{\pm}2.97$ & $69.67{\pm}2.72$ & $37.18{\pm}2.33$ & $15.76{\pm}2.65$ & $59.48{\pm}0.18$ & $9.58{\pm}2.09$ \\
    \bottomrule
  \end{tabular}
\end{table*}

\subsection{Full Seed-Level Difference Table (Tier~1)}
\label{app:full_ci}

Table~\ref{tab:ci_full} reports all six Tier~1 pairwise comparisons across all 7 benchmarks.
These are the exhaustive differences underlying the forest plot in Figure~\ref{fig:forest} and the summary in Table~\ref{tab:seed_effect}.

\begin{table*}[!htbp]
  \centering\small
  \caption{Tier~1 seed-level paired differences ($n{=}5$ seeds), all 7 benchmarks.
  Values are mean paired difference in pp.}
  \label{tab:ci_full}
  \setlength{\tabcolsep}{3pt}
  \begin{tabular}{@{}lccccccc@{}}
    \toprule
    \textbf{Comparison} & \textbf{GSM8K} & \textbf{Minerva} & \textbf{MATH} & \textbf{Olymp.} & \textbf{AIME\,'24} & \textbf{AMC\,'23} & \textbf{AIME\,'25} \\
    \midrule
    Bot$-$Rnd
      & $+2.0$ & $+2.9$ & $-0.6$ & $+1.6$ & $-1.6$ & $+0.7$ & $+0.5$ \\
    Top$-$Rnd
      & $+0.4$ & $+1.2$ & $+0.1$ & $+0.7$ & $+0.1$ & $+1.4$ & $+0.8$ \\
    Mid$-$Rnd
      & $-0.3$ & $+1.3$ & $-0.6$ & $+1.1$ & $+0.6$ & $+2.8$ & $+0.7$ \\
    Bot$-$Top
      & $+1.6$ & $+1.7$ & $-0.6$ & $+0.9$ & $-1.8$ & $-0.7$ & $-0.3$ \\
    Mid$-$Top
      & $-0.7$ & $+0.1$ & $-0.7$ & $+0.4$ & $+0.5$ & $+1.4$ & $-0.2$ \\
    Bot$-$Mid
      & $+2.3$ & $+1.6$ & $+0.0$ & $+0.5$ & $-2.2$ & $-2.1$ & $-0.2$ \\
    \bottomrule
  \end{tabular}
\end{table*}

\subsection{Tier~2 and 1.5B Seed-Level Differences}
\label{app:tier2_ci}

Table~\ref{tab:ci_tier2} reports Tier~2 pairwise comparisons ($n{=}3$ seeds).
The LIMR--Random comparison on AIME~2024 ($-1.6$\,pp) is the basis for the LIMR deficit reported in \S\ref{sec:baselines}.
The Top--LIMR row reports only the AIME comparison, as this is the only benchmark where the two differ appreciably.

\begin{table}[!htbp]
  \centering\small
  \caption{Tier~2 seed-level paired differences ($n{=}3$ seeds), 5 primary benchmarks.}
  \label{tab:ci_tier2}
  \setlength{\tabcolsep}{3.5pt}
  \resizebox{\linewidth}{!}{
  \begin{tabular}{@{}lccccc@{}}
    \toprule
    \textbf{Comparison} & \textbf{GSM8K} & \textbf{Minerva} & \textbf{MATH} & \textbf{Olymp.} & \textbf{AIME\,'24} \\
    \midrule
    LIMR$-$Rnd
      & $+1.2$ & $+0.9$ & $+1.9$ & $+1.3$ & $-1.6$ \\
    Top$-$LIMR
      & --- & --- & --- & --- & $+1.7$ \\
    Full$-$Rnd
      & $-0.3$ & $-2.5$ & $-1.1$ & $-0.4$ & $+0.5$ \\
    Full$-$LIMR
      & $-1.5$ & $-3.7$ & $-3.0$ & $-1.7$ & $+2.1$ \\
    RandH$-$Rnd
      & $+2.0$ & $+0.8$ & $+0.1$ & $+0.4$ & $-0.9$ \\
    \bottomrule
  \end{tabular}
  }
\end{table}

Table~\ref{tab:ci_15b} reports 1.5B mini-Tier~1 pairwise comparisons ($n{=}3$ seeds).
The Bot--Top comparison on MATH-500 ($-3.0$\,pp) is the basis for the cross-scale difference in \S\ref{sec:cross_scale}.
The xmod--Random comparison on GSM8K ($+1.0$\,pp) is the cross-model portability trend discussed in the same section.

\begin{table}[!htbp]
  \centering\small
  \caption{1.5B seed-level paired differences ($n{=}3$ seeds), 4 primary benchmarks.}
  \label{tab:ci_15b}
  \setlength{\tabcolsep}{3.5pt}
  \begin{tabular}{@{}lcccc@{}}
    \toprule
    \textbf{Comparison} & \textbf{MATH} & \textbf{GSM8K} & \textbf{Olymp.} & \textbf{AIME\,'24} \\
    \midrule
    Top$-$Rnd
      & $+1.8$ & $+1.1$ & $+1.2$ & $+0.4$ \\
    Bot$-$Rnd
      & $-1.2$ & $+0.1$ & $+0.4$ & $+0.8$ \\
    Mid$-$Rnd
      & $-1.1$ & $-2.1$ & $-0.7$ & $+1.0$ \\
    xmod$-$Rnd
      & $-0.5$ & $+1.0$ & $+0.1$ & $+0.2$ \\
    Bot$-$Top
      & $-3.0$ & $-1.0$ & $-0.8$ & $+0.4$ \\
    \bottomrule
  \end{tabular}
\end{table}

% ============================
\section{Per-Seed Results}
\label{app:per_seed}

\begin{figure*}[!htbp]
  \centering
  \includegraphics[width=0.95\textwidth]{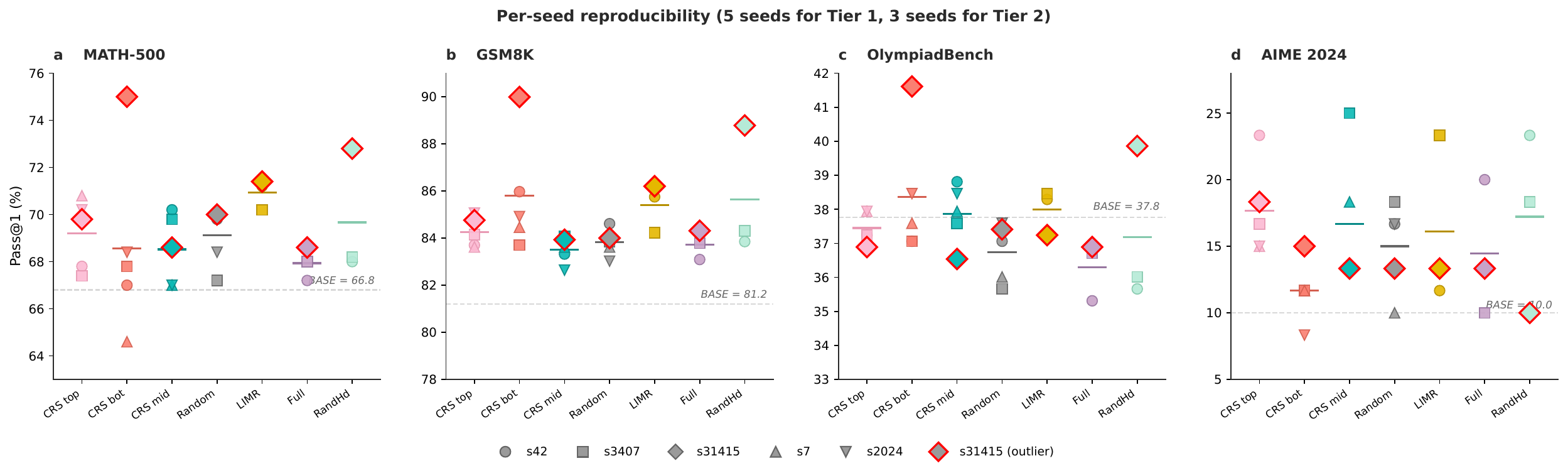}
  \caption{Per-seed accuracy for Tier~1 conditions (7B) across five primary benchmarks.
  Each dot is one (condition, seed) evaluation.
  CRS-bottom exhibits higher cross-seed variance than other conditions, particularly on MATH-500 and GSM8K.}
  \label{fig:per_seed}
\end{figure*}

Figure~\ref{fig:per_seed} visualizes the per-seed spread for Tier~1 conditions across five primary benchmarks, providing full transparency on cross-seed variance.
Tables~\ref{tab:seed_tier1}--\ref{tab:seed_15b} report the underlying per-seed accuracies.
Seeds are ordered: s42, s3407, s31415, s2024, s7 (Tier~1) or s42, s3407, s31415 (Tier~2 / 1.5B).

\begin{table}[!htbp]
  \centering\small
  \caption{Tier~1 per-seed results (5 seeds), 5 primary benchmarks.}
  \label{tab:seed_tier1}
  \setlength{\tabcolsep}{3pt}
  \resizebox{\linewidth}{!}{%
  \begin{tabular}{@{}llccccc@{}}
    \toprule
    \textbf{Condition} & \textbf{Seed} & \textbf{GSM8K} & \textbf{Minerva} & \textbf{MATH} & \textbf{Olymp.} & \textbf{AIME} \\
    \midrule
    \multirow{5}{*}{CRS-top}
      & 42    & 83.70 & 21.69 & 67.80 & 37.24 & 16.67 \\
      & 3407  & 84.15 & 18.01 & 67.40 & 37.24 & 18.54 \\
      & 31415 & 84.76 & 20.22 & 69.80 & 36.89 & 16.04 \\
      & 2024  & 85.06 & 20.96 & 70.20 & 37.94 & 17.92 \\
      & 7     & 83.62 & 23.16 & 70.80 & 37.94 & 13.96 \\
    \midrule
    \multirow{5}{*}{CRS-mid}
      & 42    & 83.32 & 19.85 & 70.20 & 38.81 & 17.92 \\
      & 3407  & 84.08 & 23.53 & 69.80 & 37.59 & 19.58 \\
      & 31415 & 83.93 & 18.75 & 68.60 & 36.54 & 17.50 \\
      & 2024  & 82.64 & 22.79 & 67.00 & 38.46 & 14.37 \\
      & 7     & 83.62 & 19.49 & 67.00 & 37.94 & 16.04 \\
    \midrule
    \multirow{5}{*}{CRS-bot}
      & 42    & 85.97 & 21.69 & 67.00 & 37.06 & 15.42 \\
      & 3407  & 83.70 & 22.79 & 67.80 & 37.06 & 15.62 \\
      & 31415 & 89.99 & 27.57 & 75.00 & 41.61 & 13.13 \\
      & 2024  & 84.91 & 20.59 & 68.40 & 38.46 & 12.71 \\
      & 7     & 84.46 & 19.85 & 64.60 & 37.59 & 17.50 \\
    \midrule
    \multirow{5}{*}{Random}
      & 42    & 84.61 & 19.85 & 69.80 & 37.06 & 17.08 \\
      & 3407  & 83.85 & 20.59 & 67.20 & 35.66 & 17.92 \\
      & 31415 & 84.00 & 20.59 & 70.00 & 37.41 & 16.04 \\
      & 2024  & 83.02 & 19.12 & 68.40 & 37.59 & 16.88 \\
      & 7     & 83.62 & 18.01 & 70.20 & 36.01 & 14.58 \\
    \bottomrule
  \end{tabular}
  }
\end{table}

\begin{table}[!htbp]
  \centering\small
  \caption{Per-seed results for LIMR, Full data, and RandHeads (3 seeds), 5 primary benchmarks.}
  \label{tab:seed_tier2}
  \setlength{\tabcolsep}{3pt}
  \begin{tabular}{@{}llccccc@{}}
    \toprule
    \textbf{Cond.} & \textbf{Seed} & \textbf{GSM} & \textbf{Min.} & \textbf{M5} & \textbf{Oly.} & \textbf{AIME} \\
    \midrule
    \multirow{3}{*}{LIMR}
      & 42    & 85.75 & 21.69 & 71.20 & 38.29 & 16.04 \\
      & 3407  & 84.23 & 18.38 & 70.20 & 38.46 & 15.83 \\
      & 31415 & 86.20 & 23.53 & 71.40 & 37.24 & 14.37 \\
    \midrule
    \multirow{3}{*}{Full}
      & 42    & 83.09 & 19.12 & 67.20 & 35.31 & 19.17 \\
      & 3407  & 83.78 & 19.12 & 68.00 & 36.71 & 16.88 \\
      & 31415 & 84.31 & 14.34 & 68.60 & 36.89 & 16.46 \\
    \midrule
    \multirow{3}{*}{RandH}
      & 42    & 83.85 & 17.28 & 68.00 & 35.66 & 17.08 \\
      & 3407  & 84.31 & 20.96 & 68.20 & 36.01 & 17.50 \\
      & 31415 & 88.78 & 23.16 & 72.80 & 39.86 & 12.71 \\
    \bottomrule
  \end{tabular}
\end{table}

\begin{table}[!htbp]
  \centering\small
  \caption{1.5B per-seed results (3 seeds), 4 primary benchmarks.}
  \label{tab:seed_15b}
  \setlength{\tabcolsep}{3pt}
  \begin{tabular}{@{}llcccc@{}}
    \toprule
    \textbf{Cond.} & \textbf{Seed} & \textbf{M5} & \textbf{GSM} & \textbf{Oly.} & \textbf{AIME} \\
    \midrule
    \multirow{3}{*}{Top}
      & 42    & 63.60 & 76.27 & 29.90 & 8.54 \\
      & 3407  & 68.80 & 82.41 & 35.14 & 10.00 \\
      & 31415 & 63.80 & 73.69 & 29.20 & 11.04 \\
    \midrule
    \multirow{3}{*}{Bot}
      & 42    & 62.40 & 78.32 & 30.59 & 11.46 \\
      & 3407  & 63.00 & 78.09 & 31.12 & 8.96 \\
      & 31415 & 61.80 & 72.93 & 30.07 & 10.21 \\
    \midrule
    \multirow{3}{*}{Rnd}
      & 42    & 64.40 & 76.88 & 29.72 & 10.00 \\
      & 3407  & 62.80 & 75.36 & 30.24 & 9.38 \\
      & 31415 & 63.60 & 76.80 & 30.77 & 8.96 \\
    \midrule
    \multirow{3}{*}{xmod}
      & 42    & 64.00 & 76.72 & 30.07 & 10.42 \\
      & 3407  & 61.80 & 76.95 & 30.42 & 10.21 \\
      & 31415 & 63.60 & 78.32 & 30.42 & 8.33 \\
    \bottomrule
  \end{tabular}
\end{table}

% ============================
\section{Supplementary Analyses}
\label{app:supp_analyses}

This section collects the analyses referenced in the main text: item-level McNemar tests, subset overlap statistics, the CRS distribution, the post-training head ablation, the hierarchical model of the condition effect, the two ends of the CRS ranking, and the construction of the difficulty-filtering subsets.

\subsection{McNemar Item-Level Tests}
\label{app:mcnemar}

For greedy-decoded benchmarks, we run McNemar's exact test on item-level binary outcomes for CRS-bottom vs.\ Random (using the median-accuracy seed).
This is an item-level statement about a single pair of runs and is complementary to the seed-level analysis of \S\ref{sec:bottom_wins}.
The GSM8K result ($p = 3.3 \times 10^{-4}$) reflects a net 38 additional problems solved by CRS-bottom that Random does not.
MATH-500 and OlympiadBench do not reach item-level significance with a single seed pair, consistent with the smaller per-condition accuracy differences on these benchmarks.

\begin{table}[!htbp]
  \centering\small
  \caption{McNemar test: CRS-bottom vs.\ Random.
  $a$: bottom correct, random wrong; $b$: reverse; ``both'': both correct.}
  \label{tab:mcnemar}
  \begin{tabular}{@{}lcccc@{}}
    \toprule
    \textbf{Benchmark} & $a$ & $b$ & \textbf{both} & $p$ \\
    \midrule
    GSM8K ($N{=}1319$) & 73 & 35 & 1081 & $3.3{\times}10^{-4}$ \\
    MATH-500 ($N{=}500$) & 16 & 23 & 329 & 0.337 \\
    OlympiadBench ($N{=}562$) & 23 & 14 & 193 & 0.188 \\
    \bottomrule
  \end{tabular}
\end{table}

\subsection{Subset Overlap Analysis}
\label{app:overlap}

Table~\ref{tab:overlap} quantifies the independence of different selection methods.
All Jaccard values are at or near chance, confirming that CRS, SR (success-rate-based), and LIMR rank problems in nearly independent orders.

\begin{table}[!htbp]
  \centering\small
  \caption{Jaccard similarity between selection subsets (both at 10\% of pool).
  Chance overlap $\approx 0.05$ for same-pool pairs, $\approx 0.16$ for cross-pool pairs.}
  \label{tab:overlap}
  \begin{tabular}{@{}lcc@{}}
    \toprule
    \textbf{Pair} & \textbf{Jaccard} & \textbf{Chance} \\
    \midrule
    CRS-top $\cap$ CRS-bottom & 0.000 & 0.053 \\
    CRS-top $\cap$ SR-top (easiest) & 0.078 & 0.053 \\
    CRS-bot $\cap$ SR-bot (hardest) & 0.055 & 0.053 \\
    Oly-CRS-top $\cap$ LIMR & 0.153 & 0.163 \\
    Oly-CRS-bot $\cap$ LIMR & 0.165 & 0.163 \\
    \bottomrule
  \end{tabular}
\end{table}

\subsection{CRS Distribution}
\label{app:crs_dist}

\begin{figure}[!htbp]
  \centering
  \includegraphics[width=\columnwidth]{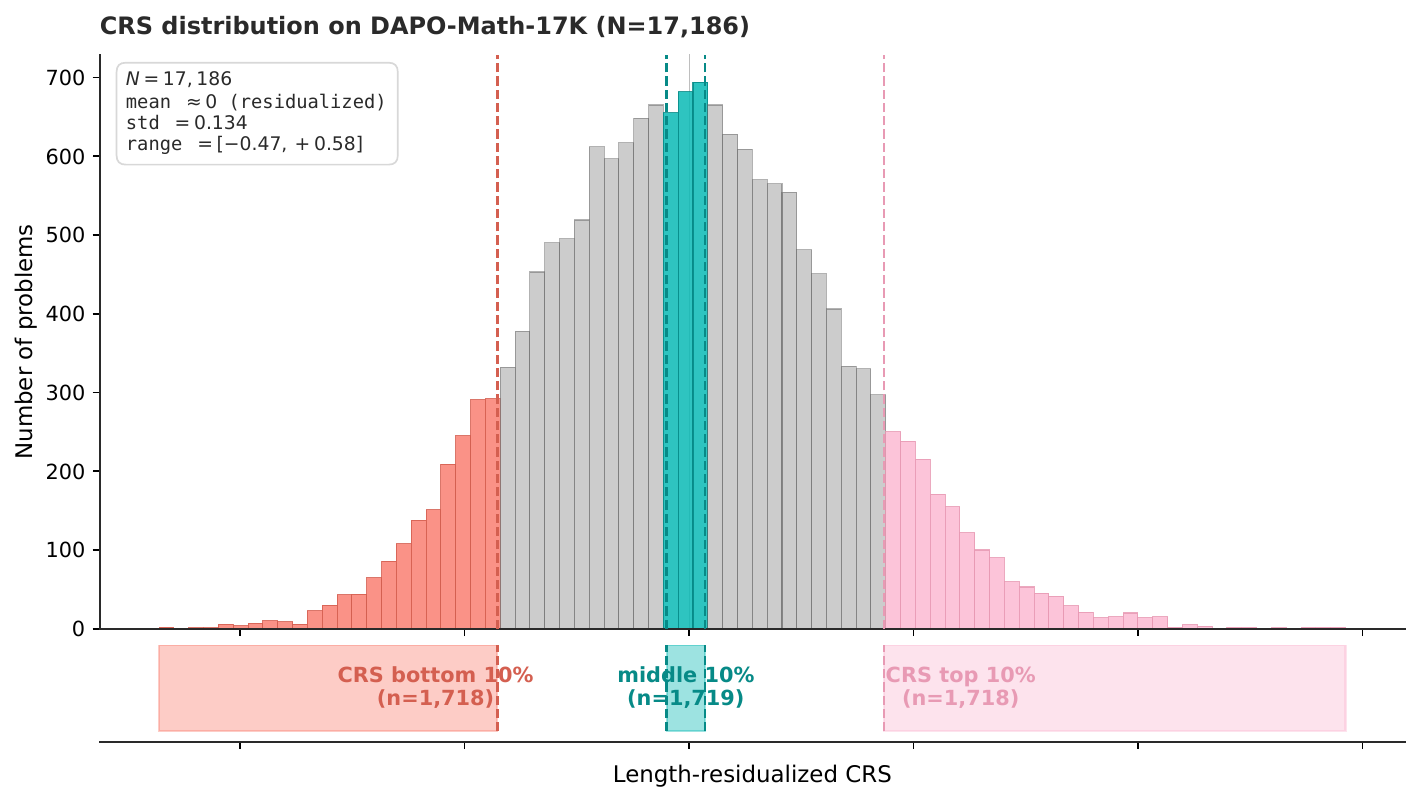}
  \caption{Distribution of length-residualized CRS ($N{=}17{,}186$).
  Colored bands: bottom-10\% (pink), middle-10\% (gray), top-10\% (teal).
  The three selection deciles occupy distinct, non-overlapping regions of the distribution.}
  \label{fig:crs_dist}
\end{figure}

Figure~\ref{fig:crs_dist} shows the distribution of length-residualized CRS across all 17,186 DAPO-17K problems.
The distribution is approximately Gaussian with mild right skew.
Colored bands indicate the three selection deciles: bottom 10\% (pink), middle 10\% (gray), and top 10\% (teal).
The decile boundaries are well-separated, confirming that the three subsets represent genuinely different regions of the CRS ranking.

\subsection{Post-Training Head Ablation}
\label{app:ablation}

We zero-ablate the 46 identified heads in the trained models (CRS-bottom and Random, median seed, following the McNemar convention of \S\ref{app:mcnemar}), against six layer-matched random 46-head control sets and in the untrained base model, with the read-out criterion fixed in advance.
DR($S$) denotes the reduction in CRS-bottom's advantage over Random when head set $S$ is zero-ablated in both trained models.

Ablating the identified heads removes the CRS-bottom advantage on every benchmark, leaving a residual advantage near zero.
DR uses the median seed, so it can slightly exceed the five-seed mean advantage (e.g.\ $+2.04$ vs.\ $+1.61$ on OlympiadBench).
On Minerva the identified set yields the largest reduction of the seven sets tested; on GSM8K and OlympiadBench it ranks third, and the seven sets are not well separated there.

The same heads are critical to the untrained base model: on GSM8K, where the three models can be compared directly, ablation lowers base accuracy from $81.20\%$ to $0.23\%$, a loss of $81.0$\,pp, against $86.0$ and $84.2$\,pp for the two trained models.
Had GRPO built new capability into these heads, ablation should harm the trained models far more than the base model; the matched losses instead indicate a pathway already present in the base model, which training uses but does not construct.
Zeroing 46 whole heads is coarse and disables mathematical generation broadly (parse-failure rates rise to $73$--$88\%$), so this establishes that the heads are critical to mathematical generation, not that they form a reasoning-specific circuit.

\begin{table}[!htbp]
  \centering\small
  \caption{Post-training zero-ablation of the 46 identified heads.}
  \label{tab:ablation}
  \begin{tabular}{@{}lccc@{}}
    \toprule
    & \textbf{Minerva} & \textbf{GSM8K} & \textbf{Olymp.} \\
    \midrule
    DR(identified 46), pp & $+1.23$ & $+1.79$ & $+2.04$ \\
    rank among 7 sets & 1\,/\,7 & 3\,/\,7 & 3\,/\,7 \\
    base acc., pre-ablation & 16.18 & 81.20 & 37.76 \\
    base acc., post-ablation & \phantom{0}0.74 & \phantom{0}0.23 & \phantom{0}0.00 \\
    \bottomrule
  \end{tabular}
\end{table}

\subsection{Hierarchical Model of the Condition Effect}
\label{app:glmm}

Seed-level pairing removes effects shared by both conditions within a seed, but not the condition-by-seed interaction.
We therefore fit a binomial GLMM, \texttt{correct $\sim$ condition + (1 + condition | seed) + (1 | problem)}, over 21,530 item-level outcomes on GSM8K, Minerva, and OlympiadBench ($1{,}319 + 272 + 562$ items, across five seeds and both conditions; MCMC, $\hat{R} = 1.002$).
The condition coefficient is $+0.256$ log-odds, with $91.3\%$ of the posterior above zero.
The seed-level slope SD is $\sigma_t = 0.372$, of the same order as the condition coefficient, so the model treats the interaction as a measured quantity rather than an assumption.

Two further checks accompany this estimate.
Removing any single seed leaves the sign of the CRS-bottom advantage unchanged on every benchmark.
The exact sign-flip permutation test over five seed composites has a two-sided floor of $0.0625$, so we report effect sizes rather than significance labels throughout.

\subsection{The Two Ends of the CRS Ranking}
\label{app:ushape}

CRS-top's point estimate exceeds Random on all three target benchmarks, so the relationship between CRS and downstream accuracy is not monotone.
Read seed by seed, the two ends are asymmetric (Table~\ref{tab:seed_effect}): CRS-bottom's advantage is positive on 13 of 15 seed-by-benchmark cells, CRS-top's on 10, and CRS-middle's on 8.
The pattern is best described as an asymmetric U---a large, seed-stable CRS-bottom effect alongside a weaker CRS-top effect not separable from CRS-middle at this seed count.

\subsection{Difficulty-Filtering Subset Construction}
\label{app:difficulty}

The two difficulty bands of \S\ref{sec:baselines} are built from the rollout-estimated success rates of \S\ref{sec:not_difficulty} ($K{=}4$ at $\tau{=}0.6$; valid estimates for 16,478 of the 17,186 pool problems).
Each band holds 1,647 problems, $10\%$ of the SR-scored pool, matching the selection ratio used for the CRS deciles.
Because $K{=}4$ quantizes SR onto five values, we select by rank rather than by threshold: \textbf{Diff-hard} takes the 1,647 lowest-SR problems with $\text{SR}>0$---at $\text{SR}{=}0$ every completion fails, the group reward has zero variance, and GRPO receives no gradient---and \textbf{Diff-easy} the 1,647 highest.
Quantization makes Diff-hard entirely $\text{SR}{=}0.25$; conditions are labelled by measured mean SR.
Jaccard overlap between these bands and the CRS deciles ranges from $0.030$ to $0.069$ across the four cross-pairings, against a chance level of ${\approx}0.051$, consistent with Table~\ref{tab:overlap}.
Difficulty filtering therefore required $4 \times 16{,}478 = 65{,}912$ generations before any training run, against a single forward pass per problem for CRS.

\begin{figure*}[!htbp]
  \centering
  \includegraphics[width=0.95\textwidth]{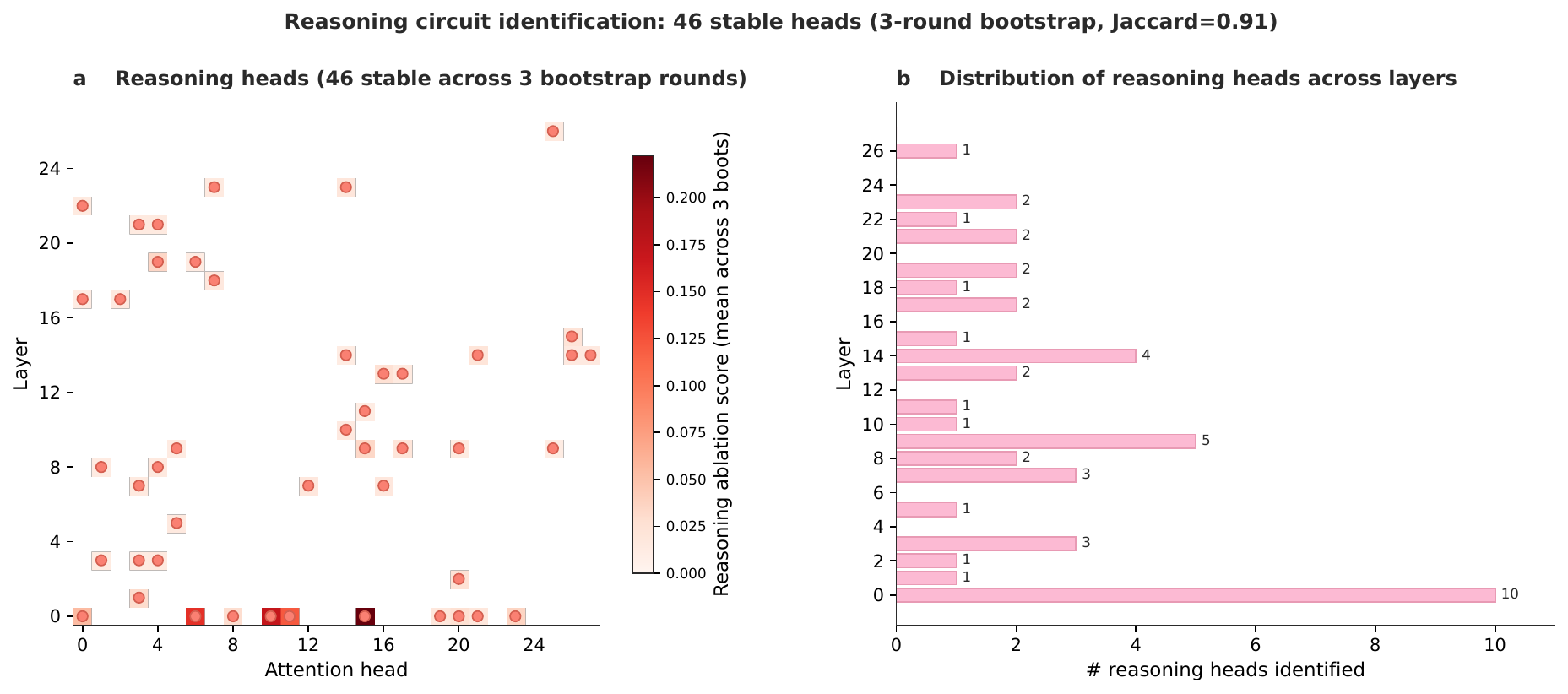}
  \caption{Bootstrap selection frequency across 784 heads ($28 \times 28$).
  Dark cells: selected in 3/3 bootstrap rounds.
  Green boxes: 46 retained heads.
  Red dashed: L0H3 (excluded global bottleneck).}
  \label{fig:circuit_id}
\end{figure*}

% ============================
\section{Circuit Identification Details}
\label{app:circuit}

Figure~\ref{fig:circuit_id} visualizes the bootstrap selection frequency across all 784 heads.
L0H3 (red dashed box), excluded due to its outlier contrastive score ($4.8\times$ the runner-up), shows extreme loss increases on \emph{both} probe types, suggesting a global information bottleneck rather than reasoning specialization.

\begin{figure*}[!htbp]
  \centering
  \includegraphics[width=0.95\textwidth]{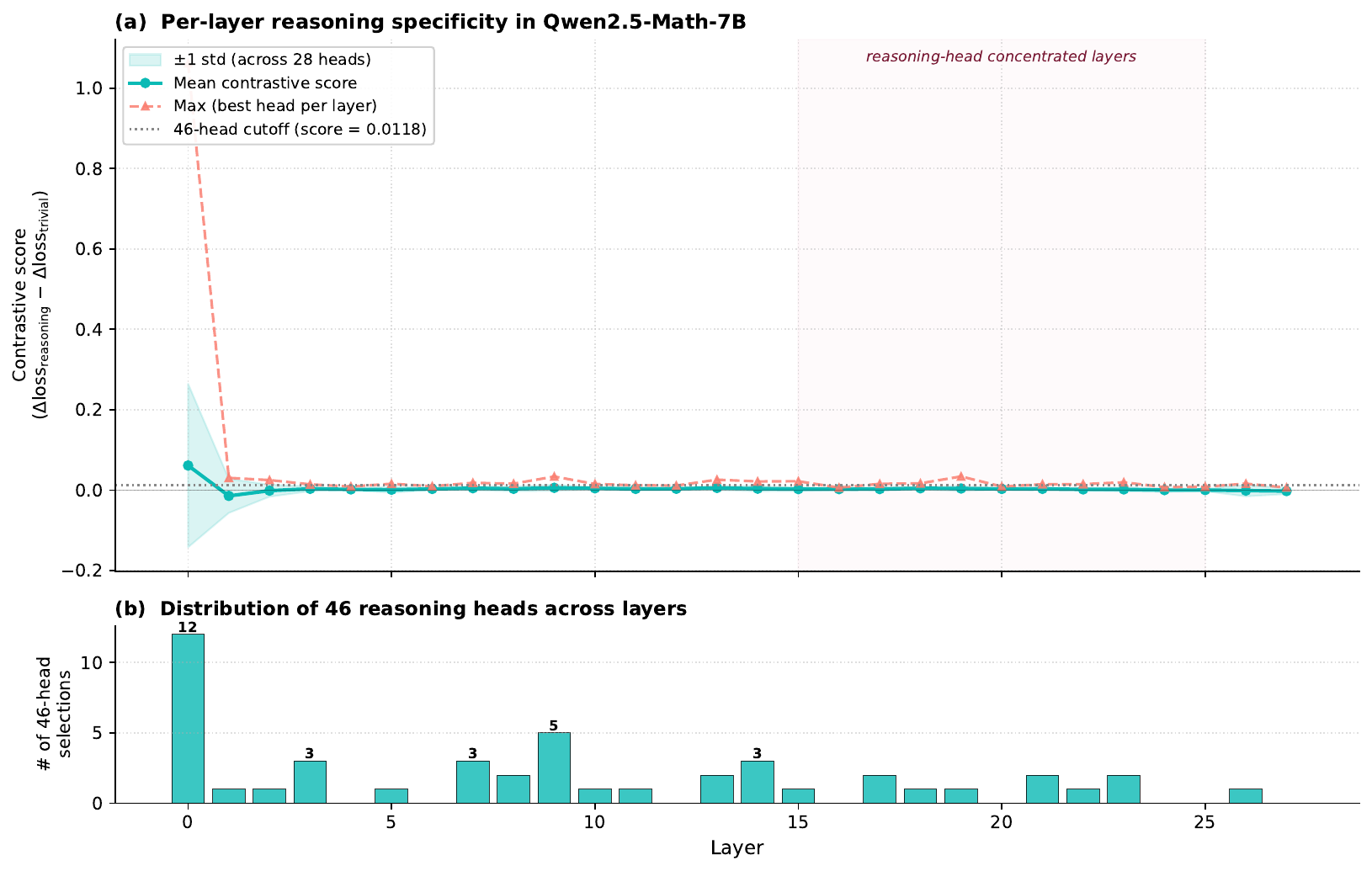}
  \caption{Per-layer contrastive scores.
  Each point is one attention head; highlighted points are the 46 selected heads.}
  \label{fig:probing}
\end{figure*}

Figure~\ref{fig:probing} shows the per-layer distribution of the 46 heads' contrastive scores.
The signal is spread across the network, with concentrations near layer~0 and in the mid-to-late layers; layers 26--27 contribute few heads, suggesting that the final layers are more involved in output formatting than in reasoning computation.

% ============================
\section{Sensitivity Analyses}
\label{app:sensitivity}

\paragraph{The contrastive weight in Eq.~\ref{eq:contrastive}.}
The weight $0.5$ is a heuristic $50\%$ discount on general-purpose head importance, fixed before any downstream experiment.
Varying $w$ over $\{0, 0.25, 0.5, 0.75, 1.0\}$ and re-deriving both the head set and the full CRS ranking (Table~\ref{tab:wgrid}), the head set and ranking are stable for $w \in [0.25, 1.0]$ (Jaccard $\geq 0.69$, Spearman $\geq 0.94$).
The $w = 0$ row supports the design: with the contrastive term switched off the head set drifts substantially (Jaccard $0.48$), evidence that the contrast against trivial-probe importance does real work.

\begin{table}[!htbp]
  \centering\small
  \caption{Sensitivity of the head set and CRS ranking to the contrastive weight $w$.}
  \label{tab:wgrid}
  \begin{tabular}{@{}lccc@{}}
    \toprule
    $w$ & \textbf{Jaccard (vs.\ .5)} & \textbf{Spearman (vs.\ .5)} & \textbf{heads} \\
    \midrule
    0.00 & 0.484 & 0.605 & 46 \\
    0.25 & 0.745 & 0.937 & 43 \\
    0.50 & 1.000 & 1.000 & 46 (paper) \\
    0.75 & 0.816 & 0.986 & 43 \\
    1.00 & 0.691 & 0.958 & 47 \\
    \bottomrule
  \end{tabular}
\end{table}

\paragraph{The residualization form.}
We tested $n$, $\sqrt{n}$, and $\log n$ on all 17,186 problems (Table~\ref{tab:residform}).
$\log n$ gives the best fit and, by construction, zero residual correlation with log-length.
The decisive quantity is downstream: bottom deciles under the three forms overlap pairwise at Jaccard $0.740$--$0.891$ (chance ${\approx}0.053$), so the trained subsets would be substantially the same under any of the three.

\begin{table}[!htbp]
  \centering\small
  \setlength{\tabcolsep}{4pt}
  \caption{Residualization forms ($N{=}17{,}186$).}
  \label{tab:residform}
  \begin{tabular}{@{}lcccc@{}}
    \toprule
    $f(n)$ & $R^2$ & $r(\text{resid}, n)$ & $r(\text{resid}, \log n)$ & \textbf{top-decile J.} \\
    \midrule
    $n$        & 0.3223 & $\phantom{-}0.000$ & 0.056 & 0.048 \\
    $\sqrt{n}$ & 0.3412 & $-0.010$ & 0.011 & 0.045 \\
    $\log n$   & 0.3436 & $\phantom{-}0.013$ & 0.000 & 0.058 \\
    \bottomrule
  \end{tabular}
\end{table}

\paragraph{Confidence intervals for Method-section correlations.}
Table~\ref{tab:methodci} reports 10,000-resample bootstrap $95\%$ CIs for the correlations quoted in \S\ref{sec:method}.
The interval on $\text{Corr}(\Delta_r, \Delta_t)$ is wide because the covariance is heavy-tailed: a few high-magnitude heads dominate it, and resamples omitting them pull the estimate toward zero---converging with the $w=0$ result above, the reasoning-specific signal is concentrated in a few heads.

\begin{table}[!htbp]
  \centering\small
  \setlength{\tabcolsep}{4pt}
  \caption{Bootstrap $95\%$ CIs for Method-section correlations.}
  \label{tab:methodci}
  \begin{tabular}{@{}lccc@{}}
    \toprule
    \textbf{Quantity} & $r$ & \textbf{95\% CI} & $n$ \\
    \midrule
    $\text{Corr}(\Delta_r, \Delta_t)$ & 0.755 & $[0.180, 0.921]$ & 784 \\
    between-round score corr. & 0.9998 & $[0.998, 0.9999]$ & 784 \\
    $\text{Corr}(\text{CRS}_{\text{raw}}, n)$ & 0.568 & $[0.557, 0.579]$ & 17,186 \\
    $\text{Corr}(\text{CRS}, n)$ & 0.013 & $[-0.002, +0.028]$ & 17,186 \\
    \bottomrule
  \end{tabular}
\end{table}

% ============================
\section{Training Curves}
\label{app:train_curves}

\begin{figure*}[!htbp]
  \centering
  \includegraphics[width=0.95\textwidth]{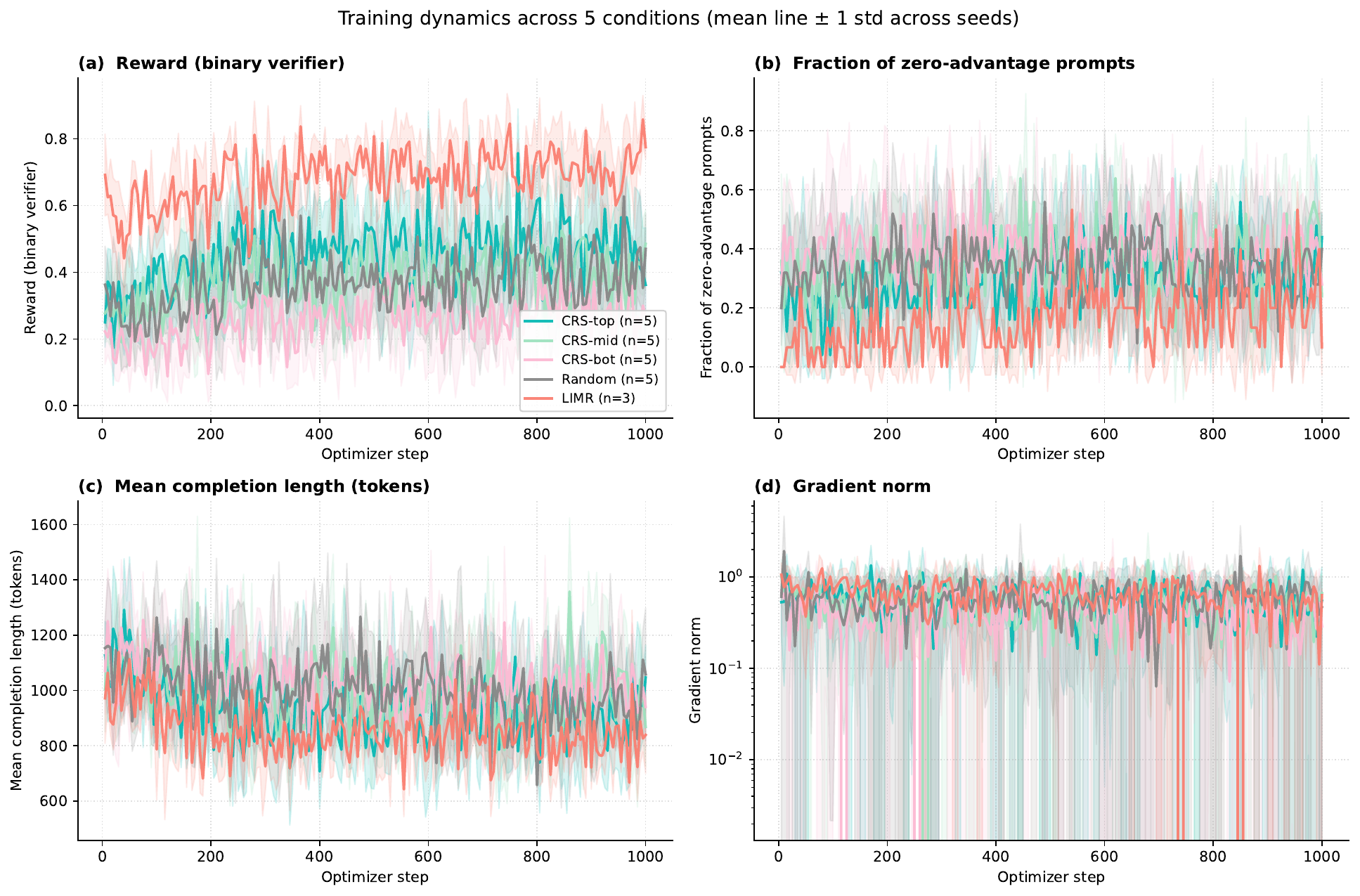}
  \caption{Training reward over 1,000 GRPO steps.
  Lines: seed-averaged reward; shaded bands: $\pm$1 std.
  LIMR saturates at the highest reward (${\sim}$0.78); CRS-bottom at the lowest (${\sim}$0.27).
  Despite this gap, CRS-bottom produces the strongest downstream generalization (Table~\ref{tab:main}).}
  \label{fig:train_curves}
\end{figure*}

Figure~\ref{fig:train_curves} shows the full training reward trajectories for the conditions plotted, with shaded bands indicating $\pm$1 standard deviation across seeds.
LIMR's reward rises steeply and saturates near 0.78 by step 600, while CRS-bottom plateaus at ${\sim}$0.27.
Despite this large gap in training reward, CRS-bottom achieves stronger downstream performance on three medium-difficulty benchmarks (\S\ref{sec:dynamics}).
The divergence in reward curves is visible from early training (step ${\sim}$200), suggesting that the training-reward--generalization decoupling is a persistent property of the data subsets, not a late-training artifact.

% ============================
\section{Implementation Details}
\label{app:impl}

\paragraph{Circuit identification.}
Three bootstrap rounds of the full 784-head contrastive ablation scan, each with independently resampled probe sets (50 reasoning + 50 trivial).
Total compute: ${\sim}$8 GPU-hours on a single A100-80GB.

\paragraph{CRS scoring.}
Single forward pass per problem on the frozen Qwen2.5-Math-7B base model.
Total compute for 17,186 DAPO problems: ${\sim}$10 minutes on a single A100.
For the 8,523 Olympiads pool: ${\sim}$5 minutes.

\paragraph{Difficulty-filtering scores.}
Rollout-estimated success rates for the difficulty baseline required $K{=}4$ generations on 16,478 problems, i.e.\ 65,912 generations in total.

\paragraph{GRPO training.}
Each run: ${\sim}$38 hours on 8$\times$A100-80GB (wall time varies with cluster preemption).
58 training runs $\times$ 38h $\times$ 8 GPUs $\approx$ 17,600 GPU-hours total.

\paragraph{Evaluation.}
Greedy benchmarks: ${\sim}$20 minutes per cell.
avg@8 benchmarks (AIME, AMC): ${\sim}$60 minutes per cell.
Total evaluation compute: ${\sim}$550 GPU-hours.

\paragraph{Overall compute budget.}
${\sim}$18,100 GPU-hours (A100-80GB equivalent), dominated by GRPO training.
CRS computation (circuit identification + scoring) accounts for less than 0.1\% of total compute.

\end{document}